\pdfoutput=1
\PassOptionsToPackage{table,xcdraw}{xcolor}
\documentclass[11pt]{article}

\usepackage[final]{acl}

\usepackage{times}
\usepackage{latexsym}

\usepackage[T1]{fontenc}
\usepackage[utf8]{inputenc}

\usepackage{microtype}

\usepackage{inconsolata}

\usepackage{amsmath} 
\usepackage{graphicx}
\usepackage{algorithm}
\usepackage{algpseudocode}
\usepackage{subcaption}
\usepackage{graphicx}
\usepackage{amsmath}
\usepackage{colortbl}
\usepackage{multirow}
\usepackage{tabularx, booktabs}
\usepackage{ulem}
\usepackage{arydshln}
\usepackage{hyperref}
\usepackage{caption}
\definecolor{Gainsboro}{rgb}{0.86, 0.86, 0.86}
\usepackage{xspace}
\usepackage[utf8]{inputenc} % allow utf-8 input
\usepackage[T1]{fontenc}    % use 8-bit T1 fonts
\usepackage{hyperref}       % hyperlinks
\usepackage{url}            % simple URL typesetting
\usepackage{booktabs}       % professional-quality tables
\usepackage{amsfonts}       % blackboard math symbols
\usepackage{nicefrac}       % compact symbols for 1/2, etc.
\usepackage{microtype}      % microtypography
\usepackage{pifont}
\usepackage[table]{xcolor}
\usepackage{colortbl}
\usepackage{amssymb}
\usepackage[breakable]{tcolorbox}
\usepackage{bbding} % 为了 \CheckmarkBold
\definecolor{green}{RGB}{36, 214, 36}
\definecolor{red}{RGB}{235, 30, 30}

\newtcolorbox[auto counter, number within=section, use counter=infoBox]{mybox}[2][]{%
    breakable,
    upperbox=visible,
    title=Prompt~\thetcbcounter: #2,
    label=#1
}

\newtcolorbox[auto counter, number within=section, use counter=blueBox]{bluebox}[2][]{%
    breakable,
    upperbox=visible,
    colback=blue!5!white,    % 背景颜色
    colframe=blue!75!black,  % 边框颜色
    title=\ourmodel Case~\thetcbcounter: #2,
    label=#1
}

\newtcolorbox[auto counter, number within=section, use counter=redBox]{redbox}[2][]{%
    breakable,
    upperbox=visible,
    colback=red!5!white,     % 背景颜色
    colframe=red!75!black,   % 边框颜色
    title=Baseline Case~\thetcbcounter: #2,
    label=#1
}

\definecolor{lightblue}{RGB}{210, 220, 250}
\newcommand{\blue}{\cellcolor{lightblue}}

\title{Improving Retrospective Language Agents via Joint Policy Gradient Optimization}

\newcommand{\ourmodel}{RetroAct\xspace}
\author{Xueyang Feng$^{1}$, Bo Lan$^{2}$, Quanyu Dai$^{3}$, Lei Wang$^{1}$, Jiakai Tang$^{1}$, \\
\textbf{Xu Chen$^{1}$\thanks{\quad Corresponding Authors.}, 
Zhenhua Dong$^{3}$, Ji-Rong Wen$^{1}$}  \\
  $^1$Gaoling School of Artificial Intelligence, Renmin University of China, Beijing, China \\
  $^2$ School of Software and Microelectronics, Peking University, Beijing, China\\
  $^3$ Huawei Noah's Ark Lab, China
  \\
    \texttt{\{xueyangfeng, xu.chen\}@ruc.edu.cn} 
    }

\begin{document}
\maketitle

\begin{abstract}
%In the latest developments in the research community, large language models (LLMs) have sparked great interest in creating autonomous agents. However, current agents built on prompt-based approaches often heavily rely on proprietary models. In contrast, agents refined through fine-tuning cannot typically self-reflect from interactions with the environment. To address these challenges, we propose a new agent architecture, named \textit{\ourmodel}, which is the first to jointly optimize both action planning and self-reflection capabilities in language agents. Specifically, we have designed a two-stage joint optimization process that includes imitation learning and reinforcement learning and developed an off-policy joint policy gradient optimization algorithm to enhance learning efficiency and stability. This process significantly improves the performance of open-source models, reduces dependency on proprietary models, and enables fine-tuned agents to learn and evolve continuousl. We conducted extensive experiments across multiple representative testing environments to validate the effectiveness of our proposed methods, confirming significant improvements in both task performance and decision-making processes. We have made all training codes and data publicly available at the following URL: \url{https://anonymous.4open.science/anonymize/RetroAct-04E8}.
In recent research advancements within the community, large language models (LLMs) have sparked great interest in creating autonomous agents. However, current prompt-based agents often heavily rely on large-scale LLMs. Meanwhile, although fine-tuning methods significantly enhance the capabilities of smaller LLMs, the fine-tuned agents often lack the potential for self-reflection and self-improvement. To address these challenges, we introduce a novel agent framework named \ourmodel, which is a framework that jointly optimizes both task-planning and self-reflective evolution capabilities in language agents. Specifically, we develop a two-stage joint optimization process that integrates imitation learning and reinforcement learning, and design an off-policy joint policy gradient optimization algorithm with imitation learning regularization to enhance the data efficiency and training stability  in agent tasks. \ourmodel significantly improves the performance of open-source models, reduces dependency on closed-source LLMs, and enables fine-tuned agents to learn and evolve continuously.
We conduct extensive experiments across various testing environments, demonstrating \ourmodel has substantial improvements in task performance and decision-making processes.
%Meanwhile, fine-tuning methods substantially enhance open-source LLMs but lack the capability for self-reflection and self-improvement.
\end{abstract}

\section{Introduction}
\label{sec:intro}
Achieving independent, autonomous agents capable of thinking, reasoning, and dynamically interacting with their environment has long been a fundamental goal for researchers in the field of artificial intelligence. In recent years, with the emergence of the powerful capabilities of Large Language Models (LLMs)~\citep{zhao2023survey}, researchers have started to utilize these models in building advanced intelligent agents~\citep{wang2024survey, yao2023react, shinn2024reflexion}. These agents demonstrate advanced capabilities in understanding complex language inputs, engaging in intricate planning~\citep{ahn2022Saycan, wei2022cot, wang2022CoT-SC, huang2022innermonologue, yao2024tot} and utilizing tools~\citep{qin2023toolllm, schick2024toolformer, shen2023hugginggpt, kong2023tptu2}.

\begin{figure}
    \centering
    \includegraphics[width=\linewidth]{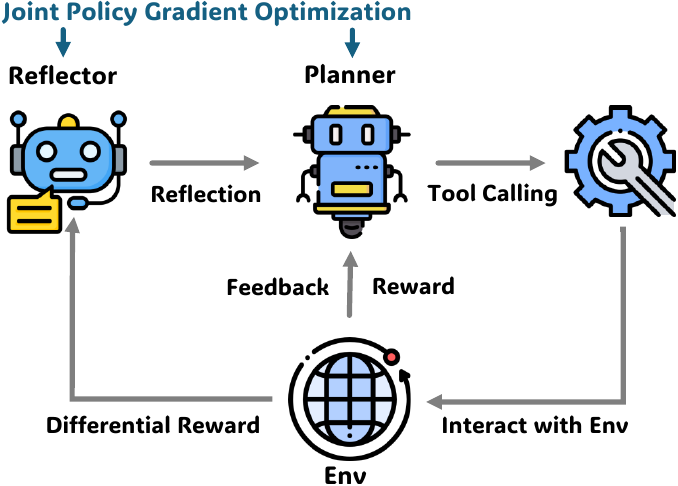}
    \caption{Overview of retrospective language agent. The planner analyzes task requirements, calls external tools, and gathers feedback. If planning fails, the reflector intervenes to adjust the strategy until the issue is resolved. Through joint strategy optimization, \ourmodel continually enhances both the planner and reflector to tackle complex tasks more effectively. }
    \label{fig:auto}
\end{figure}

\begin{table*}

  \centering
  \small
\renewcommand{\arraystretch}{1.05}
  \begin{tabularx}{\textwidth}{c *{6}{>{\centering\arraybackslash}X}}
    \toprule
    & \multicolumn{3}{c}{Planner} & \multicolumn{3}{c}{Reflector} \\
    \cmidrule(lr){2-4}\cmidrule(lr){5-7}
    Name & Prompt & IL & RL & Prompt & IL & RL \\
    \midrule
    ReAct~\citep{yao2023react} & \textcolor{cyan}{\CheckmarkBold} & \textcolor{red}{\XSolidBrush} & \textcolor{red}{\XSolidBrush} & \textcolor{red}{\XSolidBrush} & \textcolor{red}{\XSolidBrush} & \textcolor{red}{\XSolidBrush} \\
    Self-Refine~\citep{madaan2024selfrefine} & \textcolor{cyan}{\CheckmarkBold} & \textcolor{red}{\XSolidBrush} & \textcolor{red}{\XSolidBrush} & \textcolor{cyan}{\CheckmarkBold} & \textcolor{red}{\XSolidBrush} & \textcolor{red}{\XSolidBrush} \\
    Reflexion~\citep{shinn2024reflexion} & \textcolor{cyan}{\CheckmarkBold} & \textcolor{red}{\XSolidBrush} & \textcolor{red}{\XSolidBrush} & \textcolor{cyan}{\CheckmarkBold} & \textcolor{red}{\XSolidBrush} & \textcolor{red}{\XSolidBrush} \\
    Retroformer~\citep{yao2023retroformer} & \textcolor{cyan}{\CheckmarkBold} & \textcolor{red}{\XSolidBrush} & \textcolor{red}{\XSolidBrush} & \textcolor{cyan}{\CheckmarkBold} & \textcolor{cyan}{\CheckmarkBold} & \textcolor{cyan}{\CheckmarkBold} \\
    FireAct~\citep{chen2023fireact} & \textcolor{cyan}{\CheckmarkBold} & \textcolor{cyan}{\CheckmarkBold} & \textcolor{red}{\XSolidBrush} & \textcolor{red}{\XSolidBrush} & \textcolor{red}{\XSolidBrush} & \textcolor{red}{\XSolidBrush} \\
    ArCHer~\citep{xi2024training} & \textcolor{cyan}{\CheckmarkBold} & \textcolor{cyan}{\CheckmarkBold} & \textcolor{cyan}{\CheckmarkBold} & \textcolor{red}{\XSolidBrush} & \textcolor{red}{\XSolidBrush} & \textcolor{red}{\XSolidBrush} \\
    \ourmodel & \textcolor{cyan}{\CheckmarkBold} & \textcolor{cyan}{\CheckmarkBold} & \textcolor{cyan}{\CheckmarkBold} & \textcolor{cyan}{\CheckmarkBold} & \textcolor{cyan}{\CheckmarkBold} & \textcolor{cyan}{\CheckmarkBold} \\
    \bottomrule
  \end{tabularx}
  \caption{Related work on Language Agent. Prompt, IL, RL stand for Prompt-based Method, Imitation Learning-based Method, and Reinforcement Learning-based Method.}
    \label{tab:related}
\end{table*}
Since LLMs are not initially intended for agent tasks, they need to be adjusted to carry out such tasks efficiently. Currently, there are two primary adaptation paradigms: (1) \textit{Prompt-based Agent}: In this paradigm, LLMs leverage their in-context learning abilities to adapt to new environments \citep{brown2020language}. Many researchers have designed self-reflection mechanisms, allowing agents to explicitly reflect on feedback from the environment and incorporate these reflections into subsequent trials for iterative self-improvement~\citep{shinn2024reflexion, madaan2024selfrefine, yao2023retroformer}. However, these methods typically rely on large-scale LLMs, leading to substantial costs and delays. Moreover, most smaller LLMs often exhibit insufficient performance and robustness when deployed as agents~\citep{chen2023fireact}. (2) \textit{Agent Fine-tuning}. 
This paradigm primarily includes using imitation learning to fine-tune smaller LLMs with expert datasets, enabling agents to learn specific tasks from the data~\citep{zeng2023agenttuning, chen2023fireact}, as well as employing reinforcement learning to allow agents to explore and learn on training sets~\citep{zhou2024archer}. However, these fine-tuned agents depend on the knowledge acquired through fine-tuning and lack the ability to continuously learn new information and self-improve in testing environments without updating parameters. As shown in Table \ref{tab:related}, there is no research on how to simultaneously enhance an agent's task-planning abilities and self-reflection capabilities from failures, nor is there a comprehensive framework that combines IL and RL to jointly fine-tune these two capabilities of the agent. 

In this work, we propose a novel agent framework called \ourmodel, which jointly optimizes the task-planning and self-reflection capabilities of open-source LLMs. This approach eliminates the dependency on closed-source models during inference while retaining the ability for continuous reflection and evolution. Specifically, we developed a two-phase joint optimization process that integrates imitation learning and reinforcement learning. First, we use imitation learning to distill the planning and reflection capabilities of large-scale LLMs into smaller LLMs. Then, we propose an off-policy joint policy gradient optimization algorithm with imitation learning regularization to enhance data efficiency and training stability in the reinforcement learning process. During the joint optimization process, the planner and reflector can mutually facilitate each other and collectively improve the overall performance of the agent, showcasing the unique advantages of joint optimization.
%showcasing the unique advantages of joint optimization.

We conduct experiments on three representative agent tasks: Complex Reasoning (HotpotQA~\citep{yang2018hotpotqa}), Embodied Decision (ALFWorld~\citep{shridhar2020alfworld}), and Interactive Programming (InterCode~\citep{yang2023intercode}),
based on the \textit{Llama-7b} and \textit{Llama-13b}~\citep{touvron2023llama}. 
Through joint optimization, our model achieves performance improvements ranging from 22.3\% to 348.3\% on these tasks, attaining performance comparable to or exceeding that of ChatGPT~\citep{openai2022chatgpt}. Moreover, we further validate the mutual promotion between the planner and reflector. We also demonstrate that a single model can concurrently learn both planning and reflection capabilities, albeit with a slight decrease in performance. Finally, we conduct systematic ablation studies to demonstrate the importance of each module in our approach.

%To address the above issues, we introduce a new approach called \ourmodel, which jointly optimizes LLMs' planning and self-reflection capabilities. This method significantly enhances the performance of open-source models through joint policy gradient fine-tuning, thereby eliminating the dependency on proprietary models during the inference process. Specifically, we propose an agent framework that integrates self-reflection to facilitate continuous self-improvement. On this basis, we implement a two-phase fine-tuning process. First, we use imitation learning to quickly elevate the agent's planning and self-reflection capabilities to a high-performance level. Then, to further enhance performance, we design a reinforcement learning algorithm that uses off-policy policy gradient methods for joint optimization of the planner and reflector.
%our model demonstrated significant performance improvements. Specifically, compared to the prompt-based agent, \ourmodel achieved performance improvements ranging from 22.3\% to 348.3\% on these tasks. Moreover, \ourmodel exhibited continuous self-learning and evolutionary capabilities. The fine-tuned version of our 7b model achieved performance comparable to or exceeding that of ChatGPT~\citep{openai2022chatgpt}. We further verified a single 7b model can simultaneously learn planning and reflection capabilities. Additionally, we conducted systematic ablation studies to demonstrate the importance of each module in our approach.

To summarize, our contributions are shown in the following:
\begin{itemize}
\setlength{\itemsep}{0pt}
\item  %To the best of our knowledge, we are the first to 
We propose a language agent framework that jointly optimizes task-planning and self-reflective evolution capabilities. This framework enables agents built on open-source models to improve performance through fine-tuning and empowers them to continuously learn and adapt to the environment.
\item %We design an off-policy joint policy gradient optimization algorithm that enhances learning efficiency and stability.
We design an off-policy joint policy gradient algorithm with an imitation learning regularization term, which improves data efficiency and training stability.
\item We validate the effectiveness of our proposed methods through extensive experiments across multiple representative testing environments, demonstrating substantial improvements in task performance and decision-making processes.
\end{itemize}

% \section{Related Works}
% \label{sec:related_work}
% \input{sections/related_work.tex}

%\section{Preliminaries}
%\label{sec:preliminaries}
%\input{sections/preliminaries.tex}

\begin{figure*}[ht]
    \centering
    \includegraphics[width=\textwidth]{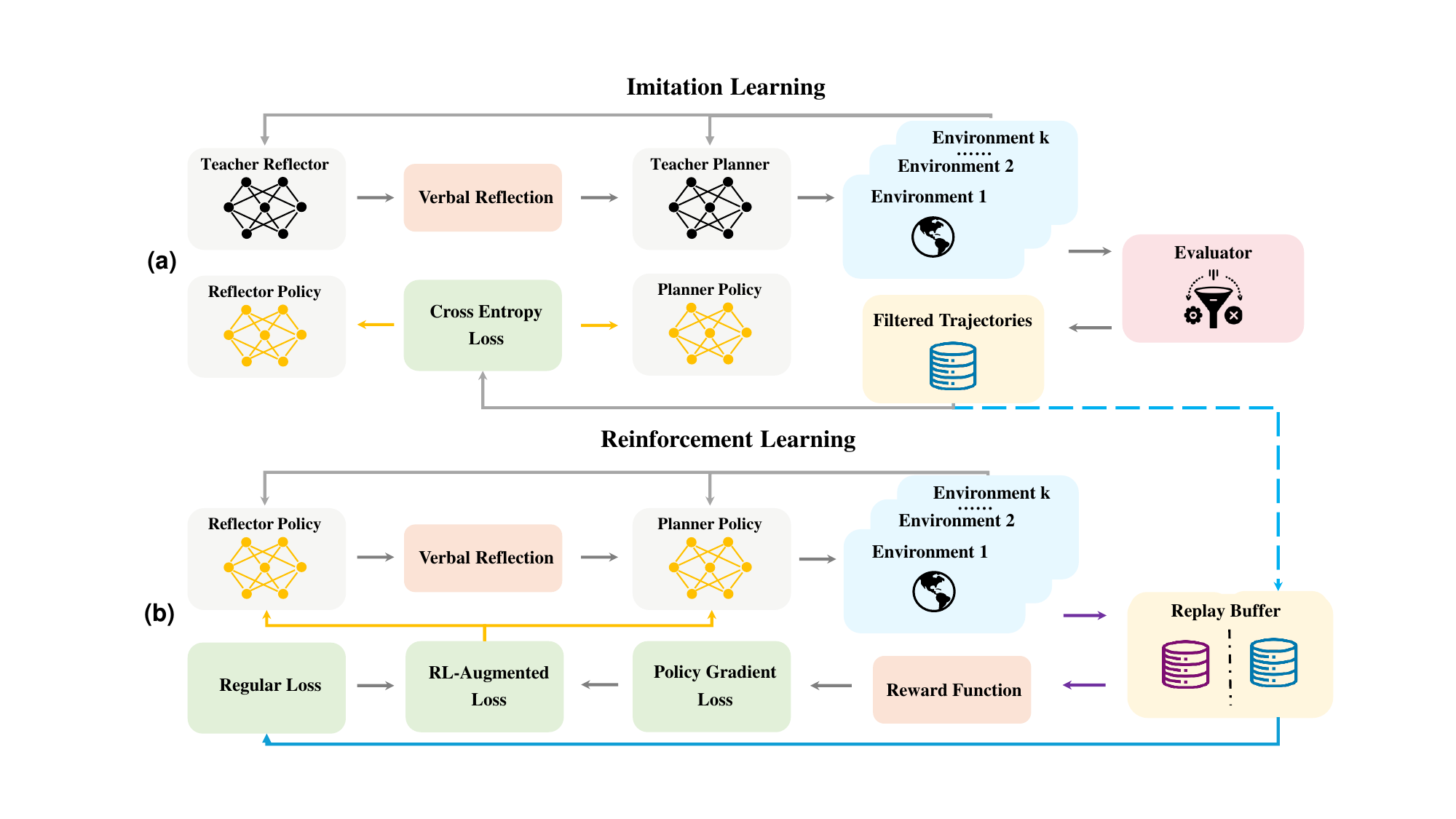}
    \caption{Schematic of Joint Policy Gradient Optimization for Retrospective Language Agent. Our approach is divided into two stages: (a) \textbf{Imitation Learning}: We use expert models to generate expert trajectories, employ evaluators to filter out these trajectories, and then use them to fine-tune the student models. (b) \textbf{Reinforcement Learning}: The planner and reflector are jointly optimized through the off-policy reinforcement learning algorithm with the imitation learning regularizer.}
    \label{fig:method}
\end{figure*}

%\section{\ourmodel: Retrospective Language Agents with Joint Policy Gradient Optimization}
\section{Approach}
\label{sec:method}

Existing methods for agent-tuning primarily rely on knowledge acquired during the fine-tuning process, lacking the ability for real-time self-reflection and self-improvement. To address these limitations, we propose \ourmodel, a novel approach that jointly optimizes LLMs' task-planning and self-reflection capabilities. Our approach significantly enhances the performance of open-source LLMs through a combination of imitation learning (IL) and reinforcement learning (RL). This eliminates the dependency on large-scale LLMs in agent tasks and retains the potential to evolve in new environments without parameter updates. Figure \ref{fig:method} illustrates the overall framework of our \ourmodel. In this section, we frame the process of LLM-based agent task-solving as a Markov Decision Process (MDP) and construct our method within this MDP framework.

%\subsection{Preliminaries: LLM-based agent for task-solving  Definitions}
\subsection{Preliminaries: LLM-based Agent for Task-Solving}
In this study, we frame LLM-based agent task-solving within the Markov Decision Process (MDP) framework, represented as the tuple $(S, A, P, O, R)$. Here, $S$ denotes the set of states, $A$ represents the available actions for the agent, $P: S \times A \times S $ defines transition probabilities between states given action, $O$ is the environmental feedback, and $R: S \times A \rightarrow R$ is the reward function, which assigns values to actions taken in different states. The LLM-based agent $\pi(a|s)$ aims to choose actions that maximize rewards. Notably, $s \in S$, $a \in A$, and $o \in O$ are all represented in natural language. Typically, $r \in R$ is sparse, with values mostly zero except for specific states, where rewards are obtained upon the conclusion of particular trajectories, such as success or failure. With a slight abuse of notation, a typical execution trajectory, consisting of $n$ steps,  is denoted as $\tau = \{s_0, a_0, o_0, \ldots, s_n, a_n, o_n, r\}$.

\subsection{Agent Architecture} 
%The \ourmodel agent architecture is similar to the Reflexion~\citep{shinn2024reflexion}, which is comprised of two language model components: a planner LLM, denoted as $\pi$, which directly interacts with the environment and generates thoughts and actions, and a reflector LLM, which generates verbal reflections to help the planner iteratively improve the plan. 
The \ourmodel agent architecture is comprised of two language model components: a planner LLM denoted as $\pi$ and a reflector LLM denoted as $\mu$. The planner directly interacts with the environment and generates thoughts and actions, while the reflector generates verbal reflections to help the planner iteratively improve the plan. It is similar to the Reflexion~\citep{shinn2024reflexion}.

\textbf{Planner Model}~
The \textit{planner model} $\pi$ resembles a policy model in reinforcement learning, employed to generate an action $a_t$ at a specified step $t$ and given state $s_t$. The state $s_t$ is textual, composed of task prompts, environmental descriptions, and historical interactions. 
The action $a_t$ comprises two distinct components: \textit{Thought} and \textit{Action}~\citep{yao2023react}. \textit{Thought} denotes the agent’s explicit thought process about the task; \textit{Action} refers to the actual interactive responses of the agent, such as utilizing tools and executing tasks. The planner can be formulated as follows:
\begin{equation}
a_t \sim \pi(\cdot|s_t)  
\end{equation}

\textbf{Reflector Model}~  
The \textit{reflector model}  $\mu$ resembles a reward model in reinforcement learning. When facing complex tasks, LLM-based agent often engages in multiple interactions with the environment to accomplish them. Tasks may fail either due to execution errors or upon encountering specific constraints. When the agent fails in the $k$-th ($0 \leq k \leq n$) trial, the unsuccessful trajectory $\tau^k$ is presented to the reflector $\mu$ to produce verbal reflections denoted as $f^k$. These verbal reflections act as semantic gradient signals to improve the planner model without parameter updates. The reflector can be formulated as:
\begin{equation}
f^k \sim \mu(\cdot|\tau^k)
\end{equation}
The initial state of the new trial is adjusted based on the initial state of the previous trial and the received feedback, i.e., $s^{k+1}_0 = s_0^{k} + f^{k}$. Consequently, the trajectory for trial $k+1$ becomes $\tau^{k+1} = \{s_0^{k+1}, a_0^{k+1}, o_0^{k+1}, \ldots, s_n^{k+1}, a_n^{k+1}, o_n^{k+1}, r^{k+1}\}$. The overall goal is to iteratively refine the policy model $\pi$ through the feedback provided by $\mu$, aiming to optimize decisions and maximize cumulative rewards across trials.

\subsection{Imitation Learning}
Despite the ability of few-shot examples and self-reflection mechanisms to adapt agents to particular tasks, these methods heavily rely on more powerful LLMs and often underperform with smaller LLMs. To overcome this issue, we first create expert datasets using large-scale LLMs. Then, we use these datasets to fine-tune the planner and reflector components of smaller LLMs. This approach enables smaller LLMs to enhance their performance by learning from the demonstrations of the larger LLMs, facilitating rapid adaptation to new tasks.

\textbf{Expert Data Collection}~ 
We utilize a powerful LLM, Mixtral-8*7b~\citep{jiang2024mixtral} with few-shot examples, serving as the teacher agents, which is denoted as $\pi_{\text{expert}}$ and $\mu_{\text{expert}}$. These teacher agents engage extensively with various environments in the training sets to generate expert trajectories for fine-tuning. We implement different rule-based evaluators for planner and reflector trajectories across multiple datasets (see Appendix \ref{app:eva} for more information). These evaluators help us filter out positive examples from the expert data to create fine-tuning datasets $D^{IL}_{\text{planner}}$ and $D^{IL}_{\text{reflector}}$.

\textbf{Imitation Learning}~
The training objective is to closely align the distribution \(\pi(a|s)\) of the planner and the distribution \(\mu(f|\tau)\) of the reflector with the expert model's action distribution. 
%Both the planner and reflector distributions are derived from the less powerful open-source model.
 This optimization objective can be expressed as:
\begin{align}
\mathcal{L}^{\text{IL}}_{\text{planner}} &= \mathbb{E}_{s \sim D^{\text{IL}}_{\text{planner}}} \left[  - \pi_{\text{expert}}(a|s) \log \pi_{\theta}(a|s) \right], \label{eq:sft_planner}\\
\mathcal{L}^{\text{IL}}_{\text{reflector}} &= \mathbb{E}_{s \sim D^{\text{IL}}_{\text{reflector}}} \left[  - \mu_{\text{expert}}(f|\tau) \log \mu_{\phi}(f|\tau) \right].  \label{eq:sft_reflector} 
\end{align}

\subsection{Joint Policy Gradient Optimization}
%Although IL has demonstrated notable efficacy, it cannot learn from broader reward signals, such as negative feedback, resulting in suboptimal results. 
%Moreover, IL simplifies the multi-step planning process into a single-step process, which may fail to capture long-term dependencies in complex tasks. 
%To this end， we reframe the language agent as a reinforcement learning agent, which can be viewed as a fusion of policy gradient optimization for both the reflector and planner policies.
Although IL has demonstrated notable efficacy, it still has several limitations. Firstly, IL relies on expert demonstrations, meaning the agent can only learn behaviors in datasets, making it difficult to surpass expert performance. Secondly, IL lacks the ability to learn from broader reward signals, such as negative feedback, which results in suboptimal outcomes when dealing with complex tasks. 
To this end, we use a joint policy gradient algorithm to simultaneously optimize both the planner and reflector. Leveraging trial-and-error and broad reward signals to learn environmental information, the planner and reflector could surpass expert demonstrations and achieve superior policies.

\textbf{Reward Shaping}~
The reward function for the planner $R_{\pi}$ is the environment-provided reward $R_{\tau}$, which is defined according to the datasets, e.g., task completion rate (more details are in Appendix \ref{app:eva}).
     Moreover, following the Retroformer~\citep {yao2023retroformer},  we design a reward function $R_{\mu}$ for the reflector, defined as the difference in environmental rewards between the trial conducted after reflection by the reflector and the last failed trial.
     %~\citep {yao2023retroformer}. %We additionally introduce a coefficient to link the rewards of the reflector to the current state's value, adjusting them as the difficulty levels of enhancing the value grows. 
 In conclusion, the reward functions of the planner and reflector in the $k$-th trial are as follows:
\begin{gather}
R_{\pi_k} = R_{\tau_{k}}, \\
R_{\mu_k} = (R_{\tau_{k+1}} - R_{\tau_{k}}).
\end{gather}

\textbf{Off-Policy Joint Policy Gradient Optimization}~
%In LLM-based agent tasks, the high inference and training costs make it challenging to rapidly iterate on training RL agents under limited computing resources~\citep{zhou2024archer}. Additionally, online RL algorithms like Policy Gradient often suffer from instability and poor performance due to low sample efficiency~\citep{song2024trialerrorexplorationbasedtrajectory}. 
In LLM-based agent tasks, online RL algorithms like Policy Gradient often suffer from poor performance due to low sample efficiency and training instability, while the training process is further hindered by high inference costs and significant latency. To address these issues, we design an off-policy joint policy gradient algorithm inspired by PPO-Clip~\citep{schulman2017proximal}.
%To address the high inference and training costs of LLMs, along with the low sample utilization in existing reinforcement learning algorithms~\citep{zhou2024archer, schulman2017proximal}, we propose an off-policy joint policy gradient algorithm. 
Specifically, we construct replay buffers to store historical trajectories and use these data in each iteration to perform off-policy optimization. %for the planner and reflector. 
The optimization objective is formulated as follows:
%
%\begin{align}
%\mathcal{L}^{\text{RL}}_{\text{planner}} &= \mathbb{E}_{s \sim D^{\text{RL}}_{\text{planner}}} \left[ - \sum_{a} w_{\pi}(s, a) R_{\pi}(s, a) \right], \\
%w_{\pi}(s, a) &= \text{Clip}\left(\frac{\pi_{\theta}(a | s)}{\pi_{\text{beh}}(a | s)}\right) \\
%\mathcal{L}^{\text{RL}}_{\text{reflector}} &= \mathbb{E}_{s \sim D^{\text{RL}}_{\text{reflector}}} \left[ - \sum_{a} w_{\mu}(\tau, f) R_{\mu}(\tau, f) \right], \\
%w_{\mu}(\tau, f) &= \text{Clip}\left(\frac{\mu_{\phi}(f | \tau)}{\mu_{\text{beh}}(f | \tau)}\right)
%\end{align}
\begin{align}
\mathcal{L}^{\text{RL}}_{\text{planner}} &= \mathbb{E}_{s \sim D^{\text{RL}}_{\text{planner}}} \left[ - \sum_{a} w_{\pi}(s, a) R_{\pi}(s, a) \right], \\
\mathcal{L}^{\text{RL}}_{\text{reflector}} &= \mathbb{E}_{s \sim D^{\text{RL}}_{\text{reflector}}} \left[ - \sum_{a} w_{\mu}(\tau, f) R_{\mu}(\tau, f) \right],
\end{align}
where \( w_{\pi} \) and \( w_{\mu} \) are importance sampling weights that adjust for the difference between the policies under parameter updates and the behavior policies, \( \pi_{\text{beh}} \) and \( \mu_{\text{beh}} \), which generate the data. These weights help mitigate the distribution shift in off-policy data~\citep{chen2021topk}. Given that importance sampling weights can introduce significant variance~\citep{schulman2017proximal}, we mitigate this by applying a clipping function, which limits the importance sampling term to the interval \(\{1-\epsilon, 1+\epsilon\}\):
\[
w_{\pi}(s, a) = \text{Clip}\left(\frac{\pi_{\theta}(a | s)}{\pi_{\text{beh}}(a | s)}, 1-\epsilon, 1+\epsilon\right)
\]
\[
w_{\mu}(\tau, f) = \text{Clip}\left(\frac{\mu_{\phi}(f | \tau)}{\mu_{\text{beh}}(f | \tau)}, 1-\epsilon, 1+\epsilon\right)
\]
Notably, we adopt a related technique~\citep{chen2024RLMEC, chen2021topk} to control the optimization speed by adjusting the gradient clipping coefficient, ensuring it remains within a reasonable range. When the optimization speed is too fast or too slow, the clipping coefficient helps regulate gradient changes, improving training stability. In contrast, standard PPO halts optimization when the discrepancy becomes too large, which can lead to the waste of a significant amount of samples in off-policy/offline scenarios. Additionally, the knowledge acquired during RL may conflict with the knowledge learned during IL, causing the model to forget previous knowledge and leading to a performance decrease. Therefore, we introduce a regularizer based on imitation learning to mitigate this effect and stabilize the training process. The resulting augmented RL loss functions for the planner and the reflector can be written as:
%which can be written as:
\begin{align}
\mathcal{L}^{\text{RL}}_{\text{planner, augmented}} &= \mathcal{L}^{\text{RL}}_{\text{planner}} + \lambda_{\pi} \mathcal{L}^{\text{IL}}_{\text{planner}}, \label{eq: rl_planner}\\
\mathcal{L}^{\text{RL}}_{\text{reflector, augmented}} &= \mathcal{L}^{\text{RL}}_{\text{reflector}} + \lambda_{\mu} \mathcal{L}^{\text{IL}}_{\text{reflector}}, \label{eq: rl_reflector}
\end{align}
where \(\lambda_{\pi}\) and \(\lambda_{\mu}\) are regularization weights that balance the influence of the RL objectives and the imitation learning objectives.

\section{Experiments}
\label{sec:experiments}
%In our study, we systematically explored the effectiveness of separately fine-tuning two models that work collaboratively to complete specific tasks. Our main experiment involved testing these models on three distinct datasets, utilizing two evaluation metrics across two different model sizes (7B and 13B). The results indicated similar performance levels for both sizes, suggesting that the scale of the model does not significantly impact the outcomes, thereby confirming the robustness of our methodology. Further, we investigated the feasibility of multitask training within a single model to determine if it could simultaneously learn and execute dual tasks. This exploration demonstrated that the single model experienced only minor performance losses in specific scenarios, supporting the viability of multitask training. Additionally, we conducted ablation studies focusing on fine-tuning individual modules—specifically, their planning and self-reflection and their distinct impacts on overall task Effectiveness. These experiments showed varied efficiencies of the modules, providing deeper insights into their functionalities and contributions. Lastly, we analyzed the effects of employing versus omitting reinforcement learning optimization techniques. By comparing the performance before and after applying these optimization strategies, we revealed their effectiveness and Effectiveness in enhancing model training processes, thus validating the performance benefits of reinforcement learning optimization.
In this section, we systematically explore the performance of our proposed method. Our experimental design revolves around four key questions: \textbf{Q1}: How does \ourmodel perform compared with the existing prompt-based and fine-tuning method? What are the benefits of joint optimization in improving the learning process of both the planner and reflector? \textbf{Q2}: How about integrating the planner and reflector within the same open-source model? Can the model effectively learn both planning and self-reflection capabilities simultaneously? \textbf{Q3}: How do the optimized planner and reflector respectively influence the behavior of \ourmodel? \textbf{Q4\&Q5}: How does the reinforcement learning process and imitation learning regularization impact \ourmodel?

\begin{table*}[!t]
\centering
\small
\renewcommand{\arraystretch}{1.2}
\setlength\tabcolsep{3pt}

\begin{tabularx}{\textwidth}{@{} >{\hsize=1.0\hsize\centering\arraybackslash}m{1.6cm} >{\hsize=1.8\hsize\centering\arraybackslash}X *{9}{>{\hsize=0.7\hsize\centering\arraybackslash}X} >{\hsize=0.9\hsize\centering\arraybackslash}X @{}}
\toprule
\multirow[m]{2}{*}{\textbf{Model}} & \multirow[m]{2}{*}{\textbf{Method}}
& \multicolumn{3}{c}{\textbf{HotpotQA}} 
& \multicolumn{3}{c}{\textbf{ALFWorld}} 
& \multicolumn{3}{c}{\textbf{InterCode}} & \multirow[m]{2}{*}{\textbf{Avg}}
\\
\cmidrule(lr){3-5} \cmidrule(lr){6-8} \cmidrule(lr){9-11}
& & \textbf{IR} & \textbf{FR} & \textbf{AR} & \textbf{IR} & \textbf{FR} & \textbf{AR} & \textbf{IR} & \textbf{FR} & \textbf{AR} &
\\
\hline 
\multirow[m]{5}{*}{\textit{Llama-7b}} 
& ReAct
& 39.7 & 39.7 & 39.7 & 8.96 & 8.96 & 8.96 & 14.24 & 14.24 & 14.24 & 21.15
\\
& Reflexion
& 39.7 & 58.49 & 54.39 & 8.96 & 21.64 & 18.28 & 14.24 & 32.16 & 28.40 & 30.14
\\
& SFT
& 55.58 & 55.58  & 55.58  & 72.39 & 72.39 & 72.39 &30.67 & 30.67 & 30.67 & 52.65
\\
& SFT+EI
& 58.22 & 58.22  & 58.22  & 68.66 & 68.66 & 68.66 & 34.86 & 34.86 & 34.86 & 53.91
\\
& SFT+RL
& \underline{60.70} & \underline{60.70} & \underline{60.70} & \underline{80.60} & \underline{80.60} & \underline{80.60} & \underline{39.42} & \underline{39.42} & \underline{39.42} & \underline{60.24}
\\
& \blue{\textbf{Ours}}
& \blue{\textbf{60.92}} & \blue{\textbf{71.51}} & \blue{\textbf{67.90}} & \blue{\textbf{82.84}} & \blue{\textbf{97.01}} & \blue{\textbf{93.28}} & \blue{\textbf{41.12}} & \blue{\textbf{54.17}} & \blue{\textbf{51.46}} & \blue{\textbf{68.91}}
\\ 
\hdashline
\multirow[m]{5}{*}{\textit{Llama-13b}} 
& ReAct
& 43.99 & 43.99 & 43.99 & 28.36 & 28.36 & 28.36 & 26.83 & 26.83 & 26.83 & 32.94
\\
& Reflexion
& 43.99 & \underline{62.39} & 59.73 & 28.36 & 46.27 & 39.70 & 26.83 & 43.00 & 38.79 & 43.01
\\
& SFT
& 58.99 & 58.99 & 58.99 & \textbf{79.10} & \underline{79.10} & \underline{79.10} & 43.17 & 43.17 & 43.17 & \underline{60.75}
\\
& SFT+EI
& \underline{60.23} & 60.23  & 60.23  & 71.64 & 71.64 & 71.64 & \textbf{47.01} & \underline{47.01} & \underline{47.01} & 59.63
\\
& SFT+RL
& \textbf{61.90} & 61.90 & \underline{61.90} & 77.61 & 77.61 & 77.61 & 42.25 & 42.25 & 42.25 & 60.59
\\
& \blue{\textbf{Ours}}
& \blue{57.99} & \blue{\textbf{70.11}} & \blue{\textbf{66.58}} & \blue{\underline{78.36}} & \blue{\textbf{91.04}} & \blue{\textbf{87.39}} & \blue{\underline{44.30}} & \blue{\textbf{61.83}} & \blue{\textbf{58.38}} & \blue{\textbf{68.11}} 
\\ 
\hdashline
\multirow[m]{2}{*}{\textit{ChatGPT}} 
& ReAct 
& 41.84 & 41.84 & 41.84 & 52.24 & 52.24 & 52.24 & 62.47 & 62.47 & 62.47 & 52.32
\\
& Reflexion
& 41.84 & 68.61 & 62.12 & 52.24 & 79.10 & 71.04 & 62.47 & 69.00 & 68.63 & 63.34
\\
\bottomrule
\end{tabularx}
\caption{Experimental results on HotpotQA, ALFWorld, InterCode. \textbf{Avg} is the average accuracy of all tasks. \textbf{IR}, \textbf{FR}, and \textbf{AR} stand for initial reward, final reward, and average reward, respectively. The best results and second best results are \textbf{bold} and \underline{underlined}, respectively.}
\label{tab:main}
\end{table*}

\subsection{Environmental Settings}
We select three representative agent environments: 
\textbf{(1)~Complex Reasoning: HotpotQA}~\citep{yang2018hotpotqa} is a multi-turn QA dataset. %The model needs to call external search tools multiple times to summarize the information obtained and answer complex questions.
% The agent needs to call external search tools multiple times to obtain the necessary information to answer complex questions.
Following ReAct~\citep{yao2023react} and ChatCoT~\citep{chen2023chatcot}, we reconstruct the HotpotQA environments. The agent needs to
call an external retriever based on SimCSE~\citep{gao2021simcse} multiple times to obtain the necessary information for answering complex questions.
%The reward depends on the F1 score, calculated by the commonality of words between the prediction and the gold standard text; 
\textbf{(2)~Embodied Decision: ALFWorld}~\citep{shridhar2020alfworld} is a text-based environment designed to simulate real-world interactions through embodied agents. In this setting, agents are tasked with executing a sequence of natural language actions informed by surrounding environment feedback to accomplish complex goals. 
%These tasks often require agents to engage in intricate planning and logical reasoning. The agent will receive a binary reward based on whether it completes the task; 
\textbf{(3)~Interactive Coding: InterCode}~\citep{yang2023intercode} is a framework for 
%designing interactive code environments to evaluate language agents capable of interactive coding. In this work, we utilize the InterCode-SQL to thoroughly assess the agent's interactive SQL query generation capability. 
evaluating language agents capable of interactive coding. In this work, we utilize the InterCode-SQL to thoroughly assess the agent's interactive SQL query generation capability. 
Moreover, We discuss in detail the selection criteria and configuration settings of the baselines and evaluation metrics in Appendix~\ref{app: exp details}.
%and employ Intersection over Union (IoU) as the reward for SQL command outputs generated by agents.

%needs to call a retriever based on SimCSE~\citep{gao2021simcse} to collect clues related to the question and answer the question. 

% It can effectively evaluate the ability of an agent to interact autonomously with its surrounding environment and make independent decisions.

% \subsection{Complex Reasoning: HotpotQA}
% HotpotQA~\citep{yang2018hotpotqa} is a multi-turn QA dataset. The model needs to call external search tools multiple times to summarize the information obtained and answer complex questions.
% Following ReAct~\citep{yao2023react} and ChatCot~\citep{chen2023chatcot}, we reconstructed the hotpotqa environment. The agent will be asked to call a retriever based on simcse~\citep{gao2021simcse} to collect clues related to the question and answer the question.

\subsection{Main Experiment on Multi-Agent (Q1)}

In this section, we implement the planner and reflector as two different LLMs, thereby instantiating \ourmodel as a multi-agent framework. We then compare it with baseline agents based on prompts and fine-tuning, respectively. Comparison of \ourmodel agents with other baseline agents across three environments and two base LLMs are shown in Table \ref{tab:main}. Overall, our method demonstrates significant advantages compared to baseline methods. Additionally, there is mutual facilitation between the reflector and the planner during the joint optimization process.

\textbf{Baseline Analysis}~
(1) For prompt-based methods, \textit{ChatGPT} agents outperform \textit{Llama} agents in all environments, demonstrating that agents based on open-source LLMs perform significantly worse than those based on closed-source LLMs, indicating that current open-source LLMs are not yet sufficient for handling complex agent tasks; (2) Reflexion-based agents significantly improve both average and final rewards, even with the same initial reward. Fine-tuning methods show clear advantages over prompt-based methods, underscoring the potential of self-reflection to enhance agent performance. However, current fine-tuning approaches rarely consider the joint optimization of task planning and self-reflection capabilities; (3) Benefiting from self-exploration, EI and RL methods, particularly in the 7b model, achieve better performance than SFT, with broader reward signals making RL algorithms more effective in agent tasks. Additionally, we demonstrate in the Appendix \ref{app:ppo} our specifically designed off-policy RL is better suited for agent tasks than the standard PPO algorithm.

% For prompt-based methods, \textit{ChatGPT}-based agents outperform \textit{Llama}-based agents, particularly in ALFWorld and InterCode, demonstrating \textit{ChatGPT}'s superior performance. Furthermore, Reflexion-based agents significantly improve average and final rewards, even with the same initial reward. Fine-tuning methods exhibit clear advantages over prompt-based methods. Moreover, RL shows better performance than SFT and EI when fine-tuning the planner alone.

% In summary, our baseline analysis reveals the following findings: (1) Agents based on open-source LLMs perform significantly worse than agents based on closed-source LLMs, indicating that current open-source LLMs are inadequate for handling agent tasks. (2) Self-reflection can significantly enhance agent performance, but current fine-tuning methods rarely consider joint optimization of task planning and self-reflection capabilities; (3) Compared to SFT and EI, our specifically designed off-policy RL methods can further improve performance.

\begin{figure*}[t]
    \centering
    \includegraphics[width=0.98\textwidth]{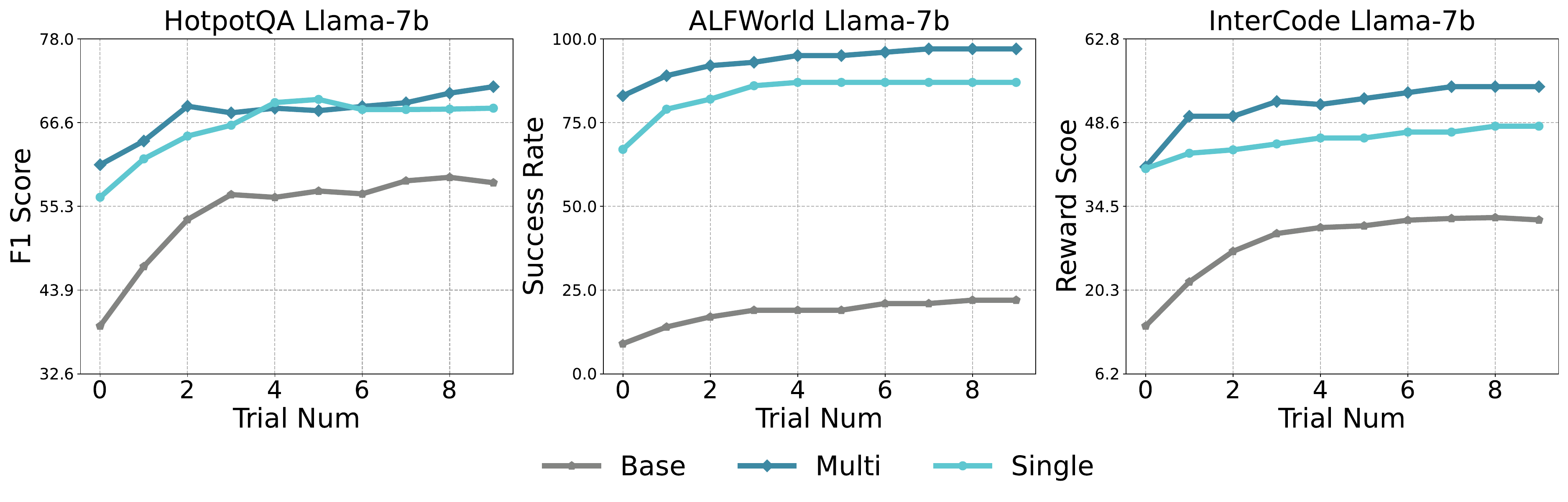}
    \caption{Multi-Agent vs Single Agent (Q2)}
     \label{fig:q2}
\end{figure*}

\begin{figure*}[t]
    \centering
    \includegraphics[width=\textwidth]{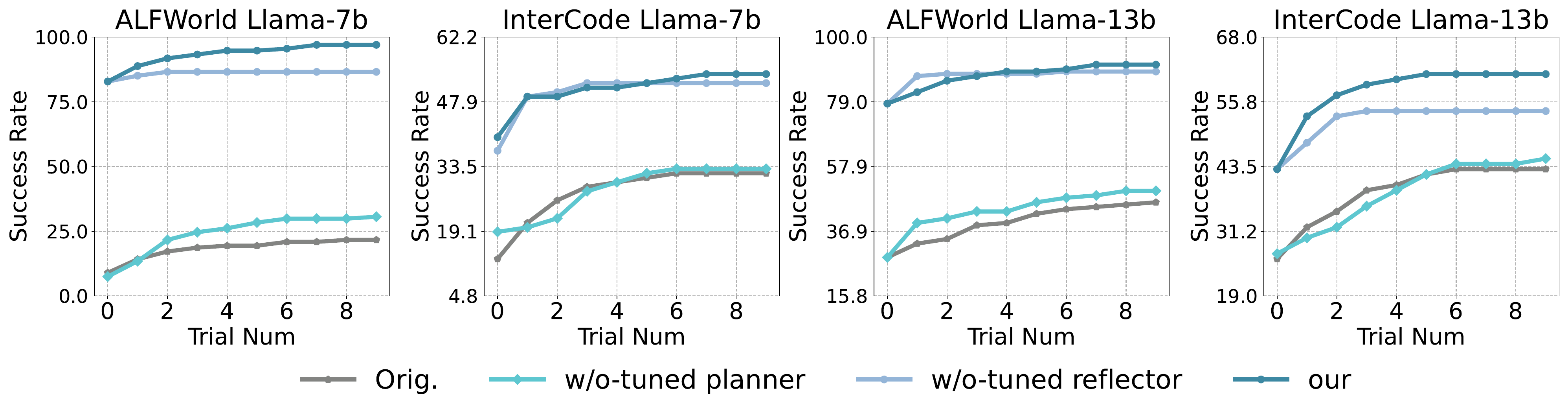}
    \caption{Effectiveness of Optimized Planner and Reflector (Q3)}
     \label{fig:q3}
\end{figure*}

\textbf{RetroAct Result}~Compared to prompt-based baselines, our approach significantly enhances the agent performance based on the same size LLMs. More notably, \ourmodel based on Llama-7b exceeds the reflexion agent based on ChatGPT by an average of 8\%.
The sustained superior performance of \ourmodel demonstrates its effectiveness in enhancing task planning and self-reflection capabilities through the knowledge gained from fine-tuning. 
Compared to the best fine-tuning baseline methods, \ourmodel achieves a 13.4\% performance improvement on average. Specifically, existing fine-tuning baseline methods do not include a reflector and cannot continuously self-improve. As a result, after multiple iterations of trials and reflections, \ourmodel shows significant advantages in both final rewards and average rewards. Furthermore, even in terms of initial rewards, \ourmodel slightly outperforms the RL baseline that optimizes the planner alone. This result underscores the \textbf{mutual facilitation between the reflector and the planner during the joint optimization process}. 
Interestingly, our approach brings more significant improvements to lower-performing base LLMs. This could be attributed to the fact that larger base models have already developed more advanced planning and reflection abilities during the pre-training phase. As a result, larger models derive less benefit from further training through supervised fine-tuning and reinforcement learning with expert or exploratory samples. This phenomenon is similar to what was observed in \citep{yuan2023scalingrelationshiplearningmathematical}.

\subsection{Comparison of Multi-Agent and Single-Agent (Q2)}

In the last experiment, we validate the effectiveness of instantiating \ourmodel as a multi-agent framework. In this section, we will further explore whether a single agent can simultaneously learn task-planning and self-reflection through fine-tuning. The results in Figure \ref{fig:q2} show that in HotpotQA, the single agent has almost no performance loss. In ALFWorld and InterCode, although our method is still effective, it has about 10\% performance loss compared to the multi-agent method. We posit that these findings are due to the difference in \textbf{trajectories consistency} between the planner and reflector. In HotpotQA, there is a high similarity between the planning and reflection tasks, as both involve standard natural language processing tasks. 
Consequently, the single agent is capable of effectively managing these tasks simultaneously. 
However, %a different phenomenon is observed in ALFWorld and InterCode. 
the ALFWorld and InterCode environments exhibit significant differences in task types: the planner in ALFWorld primarily utilizes specific natural language instruction sets, whereas the planner in InterCode involves SQL commands (we provide some trajectories in the appendix \ref{app: case}, \ref{app: prompt}). Meanwhile, reflection tasks in both environments are executed using conventional natural language. Compared to the multi-agent, the single-agent may lead to specific knowledge conflicts when learning task planning and self-reflection, resulting in decreased performance.

\subsection{Effectiveness of Optimized Planner and Reflector (Q3)}
\label{sec:q3}
In this section, we explore how the optimized planner and reflector respectively influence the behavior of \ourmodel. %Specifically, we separately removed the optimization from the planner and the reflector to study their individual impacts. 
Specifically, we separately removed the optimization from the planner and the reflector, which allowed us to study the interaction between the optimized planner and the unoptimized reflector, and vice versa. The experimental results are shown in Figure \ref{fig:q3}. Overall, removing the optimization from either the planner or the reflector negatively impacts performance. If the planner is not optimized, the agent’s performance in the first trial will degrade to the base model level. Although the optimized reflector can generally help the planner improve through trial and error, this initial performance loss is significant and cannot be recovered through self-reflection. When the reflector is not optimized, the well-optimized planner performs well in the first trial. However, due to the reflector not being sufficiently compelling, the potential for performance improvement through self-reflection is less than with our jointly optimized agent. 
%In this section, we explore how the jointly optimized planner and reflector affect the agent's specific performance. 

\begin{figure}[t]
    \centering
    \includegraphics[width=\columnwidth]{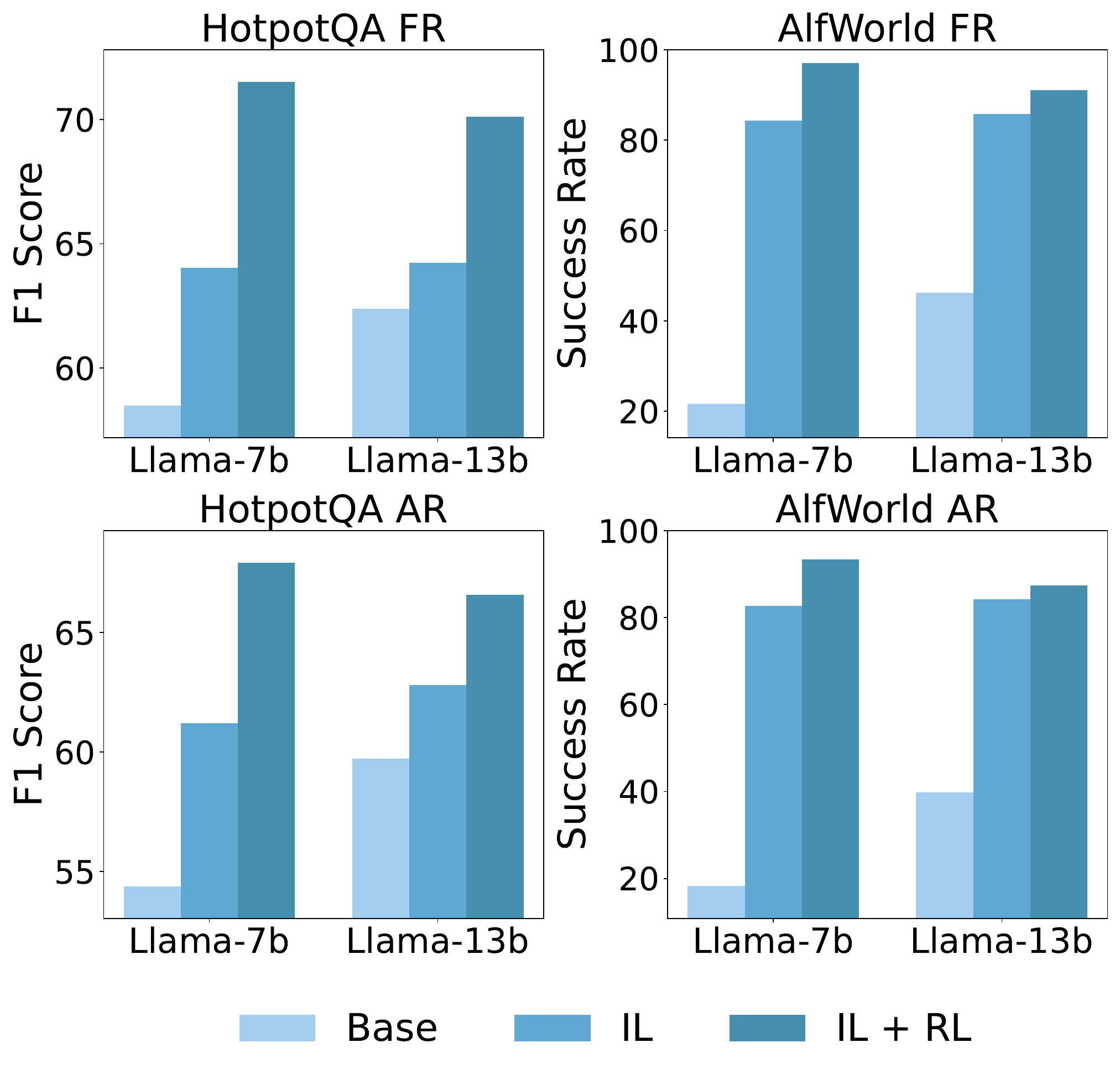}
    \caption{Effectiveness of Reinforcement Learning (Q4)}
     \label{fig:q4}
\end{figure}

Notably, compared to the base model, optimizing the planner alone is more efficient than optimizing the reflector. We attribute this to two main reasons. First, the optimization of the planner is a direct optimization of the tasks, making it more straightforward and effective. In contrast, the optimization of the reflector is essentially tuning the planner's prompts, which is less advantageous compared to directly updating the parameters. Second, there is an imbalance in the amount of data available for the planner and the reflector (we provide a comparison of effective data amounts for imitation learning in the Appendix \ref{app: exp details}), leading to the planner often being better trained.

Therefore, we conclude that for smaller LLMs, simultaneously optimizing the planner and the reflector is optimal. Additionally, optimizing the planner alone is more effective than optimizing the reflector alone. Optimizing the reflector alone is preferable when the planner cannot be optimized or the cost is too high. This finding supports the claims made in previous studies~\citep{chen2023fireact, yao2023retroformer}. To more specifically describe how the optimized planner and reflector work together to better complete tasks, we provide several cases in Appendix \ref{app: case}.

\begin{table}[t]
    \centering
    \setlength\tabcolsep{1.9pt}
    \begin{tabular}{lccc}
    \toprule
         $\lambda$ & Initial Reward & Final Reward & Average Reward\\
         \midrule
        0.0 & 52.72 & 67.71 & 64.15 \\
        1.0 & 56.52 & 68.60 & 66.24 \\
        2.0 & 55.41 & 63.19 & 62.67 \\
    \bottomrule
    \end{tabular}
    \caption{Hyperparameter Analysis on $\lambda$.}
    \label{tab:hyperparameter}
\end{table}

\begin{figure}[t]
    \centering
    \includegraphics[width=0.95\columnwidth]{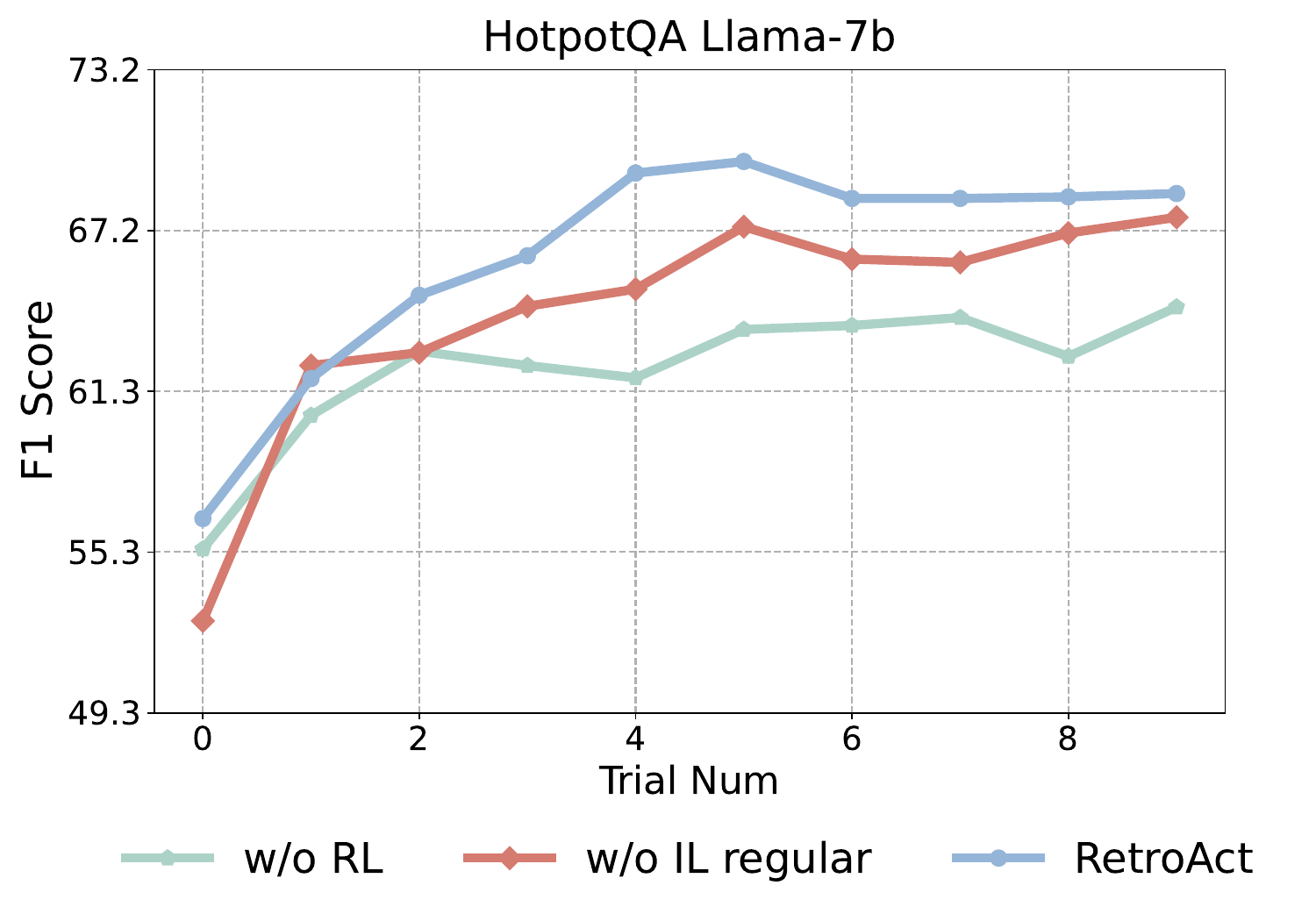}
    \caption{Effectiveness of Imitation Learning Regularization (Q5)}
    \label{fig:q5}
\end{figure}

\subsection{Effectiveness of Reinforcement Learning and Regularization (Q4\&Q5)}
%In this section, we carefully explore how our proposed off-policy joint reinforcement learning impact \ourmodel. 
 %We comprehensively evaluated and compared multi-agent, single-agent, and different model sizes, data sets, and metrics. 
 %Figure \ref{fig:q4} shows the results of our method on multi-agent and 7b models; more results are in the appendix \ref{app: abrl}. 
In this section, we explore how each component of our proposed off-policy joint reinforcement learning impacts the overall performance of \ourmodel.

\textbf{Effectiveness of RL phase}~Overall, as shown in Figure \ref{fig:q4}, we find that our RL algorithm outperformed IL in three different environments. We attribute these performance improvements to the agent's ability in RL to discover optimal behaviors beyond the constraints of the imitation learning dataset through exploration. 
Additionally, the performance improvements are more pronounced in HotpotQA and InterCode, likely because these datasets provide broader reward signals, whereas the binary reward signals in ALFWorld limit the effectiveness of RL.
%Additionally, the performance gains are more significant in HotpotQA and InterCode, likely due to the broader reward signals in these datasets, which enhance the agent's learning process. In contrast, ALFWorld's binary reward signals limit the effectiveness of RL. 
These findings underscore the importance of reward signal design in RL, suggesting that broader reward signals can better aid the agent in learning environmental information.

\textbf{Effectiveness of IL Regularization}~By setting $\lambda$ to 0, we remove the imitation learning regularization from the RL phase. As shown in Figure \ref{fig:q5}, eliminating this regularization objective led to some performance degradation, which demonstrates the role of regularization to retain the knowledge acquired during the imitation learning phase. Moreover, using the RL objective independently still resulted in acceptable performance, confirming the effectiveness of the RL objective.
Additionally, we analyze the effect of the hyperparameter \(\lambda\), which controls the weight of the regularization term, under different values. As shown in Table~\ref{tab:hyperparameter}, we find that setting the hyperparameter around 1.0 yields the best performance. When the hyperparameter value is too small, conflicts arise between the knowledge learned during the reinforcement learning (RL) phase and the knowledge acquired during the imitation learning (IL) phase, leading to performance degradation. On the other hand, when the hyperparameter value is too large, IL regularization dominates, causing the model to be overly constrained by the IL rules, thereby limiting the benefits of RL and reducing overall performance.

 %Overall, as shown in Figure \ref{fig:q4}, we find that our RL algorithm significantly improves performance over IL across three different environments. We attribute these performance gains to several vital factors. Firstly, RL allows the agent to learn from a broader range of feedback signals, enabling the agent to refine its strategies more effectively. Secondly, through trial and error, RL agents facilitate the discovery of optimal behaviors that surpass the limitations of imitation learning datasets. This is particularly beneficial in complex and dynamic environments. Lastly, the fusion of policy gradient optimization for both the reflector policy and the planner policy within our RL framework creates a mutually reinforcing learning process, allowing the agent to simultaneously enhance its planning and self-reflection capabilities to the optimal state.
 %We attribute these performance gains to the reinforcement learning's ability to allow iterative improvements and self-exploration, which synergistically interacts with the agent's self-reflection mechanism. Specifically, the self-reflection mechanism enables the agents to discover more valuable trajectories, facilitating the policy gradient learning process. Moreover, the model updates through policy gradients further enhance the agent's ability for self-reflection.

\section{Conclusion}
\label{sec:conclusion}
%In this work, we introduce \ourmodel, a agent framework that utilizes imitation learning and off-policy reinforcement learning to jointly optimize the task-planning and self-reflective capabilities of open-source LLMs. \ourmodel significantly enhances the performance of these models, reducing the dependence on closed-source LLMs in agent tasks. 
%We conduct extensive experiments across various agent environments and demonstrate that \ourmodel provides substantial improvements over baselines. Future work may involve designing more complex RL systems, such as training reward models for the planner and reflector to provide more fine-grained reward signals.

In this work, we introduce \ourmodel, an agent framework that utilizes imitation learning and off-policy reinforcement learning to jointly optimize the task-planning and self-reflective capabilities of open-source LLMs. \ourmodel significantly enhances the performance of these models, reducing the reliance on closed-source LLMs in agent tasks. We conduct extensive experiments across various agent environments, demonstrating the substantial improvements over existing baselines. Future work may involve designing more complex reinforcement learning systems, such as training reward models for the planner and reflector to provide more fine-grained reward signals.
%Additionally, it enables fine-tuned agents to continuously learn and evolve in new tasks. 
%We believe that our work has opened the door to more exciting follow-up research on the joint optimization of agent planning and reflection. Future work may involve designing more complex RL systems, such as training reward models for the planner and reflector to provide more fine-grained reward signals. Moreover, exploring how to prevent knowledge conflicts when a single agent learns both planning and reflection abilities is a promising avenue for future research.

\newpage

\section{Ethical Considerations and Limitations}
\label{app: impact}
\textbf{Ethical Considerations }~Our work leverages the powerful capabilities of Large Language Models (LLMs) to build advanced intelligent agents. While the potential benefits of these advancements are substantial, it is crucial to consider their broader impact and ethical implications. We summarize the broader impact and ethics statement of our research as follows:

\begin{itemize}
\item \textbf{Bias and Fairness}: We utilize reinforcement learning to enable open-source LLMs to learn and adapt to domain-specific agent tasks. If there are biases and discrimination in specific tasks, the agent may exacerbate these issues. Therefore, ensuring fairness in the application field is vitally important.
\item \textbf{Safety}: Despite significant efforts to align LLMs with safety standards, their deployment as agents raises additional safety concerns. Agents must avoid invoking harmful tools when interacting with external systems. Implementing constrained reinforcement learning can help ensure that agents do not engage in harmful actions, thereby enhancing their safety and reliability.
\item \textbf{Data Security}: Although we have made every effort to review and verify the data we release, some security concerns may still remain. Ensuring the integrity and security of the data is crucial to prevent potential misuse or vulnerabilities.
\end{itemize}
\label{app: limitation}
\textbf{Limitation}~Our current approach is the reliance on rewards directly provided by the environment. While this method is straightforward and effective to a certain extent, it may not always provide the most granular and informative feedback necessary for optimal performance. A more sophisticated approach would involve training separate reward models for the planner and the reflector, which could offer more detailed and tailored reward information for each component. This could potentially enhance the agent's ability to fine-tune its actions and reflections, leading to better overall performance and adaptability in complex tasks. We encourage future work to explore the development and integration of these reward models to further improve the effectiveness of our proposed approach.

\section*{Acknowledgement}
This work is supported in part by National Natural Science Foundation of China (No. 62422215 and No. 62472427), Beijing Outstanding Young Scientist Program NO.BJJWZYJH012019100020098, Intelligent Social Governance Platform, Major Innovation \& Planning Interdisciplinary Platform for the “DoubleFirst Class” Initiative, Renmin University of China, Public Computing Cloud, Renmin University of China, fund for building world-class universities (disciplines) of Renmin University of China, Intelligent Social Governance Platform, Huawei Innovation Research Programs. We gratefully acknowledge the support from Mindspore\footnote{\url{https://www.mindspore.cn}}, CANN(Compute Architecture for Neural Networks) and Ascend AI Processor used for this research.

\bibliography{anthology,custom}
\newpage

\appendix

\clearpage

\newpage
\appendix

\onecolumn
\tableofcontents

\newpage
\clearpage
\twocolumn
\section{Algorithm}
\definecolor{darkgreen}{rgb}{0, 0.5, 0}
\begin{algorithm}[h]
\caption{Practical Framework}
\label{alg:retroact_detail}
\begin{algorithmic}[1]
\State Initialize parameters $\theta, \phi, \lambda, \alpha$ % Describe parameters
\State Initialize replay buffer $D^{\text{IL}}_{\text{planner}}, D^{\text{IL}}_{\text{reflector}}, D^{\text{RL}}_{\text{planner}}, D^{\text{RL}}_{\text{reflector}}$ 
\State \textcolor{darkgreen}{\#\# IL Data Collection}
\State Initialize Expert Language Agent $\pi_{\text{expert}}, \mu_{\text{expert}}$
\For{each environment in trainset}
    \For{each trial}
        \State Update the initial state
        \State Generate $\tau^k = \{s_0^k, a_0^k, o_0^k, \dots, s_n^k, a_n^k, o_n^k\}$  using $\pi_{\text{expert}}$ 
        \State Generate verbal reward $f^k \sim \mu_{\text{expert}}(\cdot|\tau^k)$
    \EndFor
    \State Filter expert trajectories using evaluators and add to $D^{\text{IL}}_{\text{planner}}, D^{\text{IL}}_{\text{reflector}}$
\EndFor
\State \textcolor{darkgreen}{\#\# IL training}
\State Supervised Fine-tuning of $\pi_\theta$ and $\mu_\phi$ using $D^{\text{IL}}_{\text{planner}}$ and $D^{\text{IL}}_{\text{reflector}}$, respectively (Eq. \ref{eq:sft_planner} and \ref{eq:sft_reflector})
\For{each iteration}
    \State \textcolor{darkgreen}{\#\# RL exploration}
    \For{each environment in trainset}
        \For{each trial}
            \State Update the initial state 
            \State Generate $\tau^k = \{s_0^k, a_0^k, o_0^k, \dots, s_n^k, a_n^k, o_n^k\}$ using $\pi_{\theta}$, add to replay buffer $D^{\text{RL}}_{\text{planner}}$
            \State Generate verbal reward $f^k \sim \mu_{\phi}(\cdot|\tau^k)$, add to replay buffer $D^{\text{RL}}_{\text{reflector}}$
        \EndFor
    \EndFor
    \State \textcolor{darkgreen}{\#\# RL training}
    \State Update $\theta$ and $\phi$ using off-policy policy gradient optimization (Eq. \ref{eq: rl_planner} and \ref{eq: rl_reflector})
\EndFor

\end{algorithmic}
\end{algorithm}

The algorithm for RetroAct method  is shown in Algorithm \ref{alg:retroact_detail}. 

\section{Related Works}
\label{sec:related_work}
%\subsection{Retrospective LLM-agent}
%In recent years, with the emergence of the powerful capabilities of Large Language Models (LLMs), researchers have started to utilize a series of few-shot prompts in building advanced intelligent agents. They can be delved into the following two distinct types of planning modules.  

%\textbf{Planning without Feedback}~
 %Where only a single attempt is available, the agent must rely on its in-built knowledge, reasoning capabilities, and any available data to analyze situations and predict outcomes. CoT~\citep{wei2022cot} enables LLM agents to break down complex tasks into manageable subtasks, thereby solving the problem step by step. A series of research methodologies~\citep{wang2022CoT-SC,yao2024tot,qin2023DSFDT,ahn2022Saycan} generates multiple solution paths using specific methods and employs particular rules to select the final answer. Inspired by how humans often rely on tools when facing complex tasks, another series of works ~\citep{yang2024react,qin2023toolllm,schick2024toolformer,cai2023latm,shen2024hugginggpt}have utilized tools to enhance the planning abilities of LLM-agents. ReAct ~\citep{yang2024react} separates the agent's thinking and action steps, and in the action steps, the agent can choose to use external tools to obtain additional information.

\textbf{Retrospective language agent}~
% has become a common design paradigm. 
Benefitting from the LLMs' in-context learning capabilities, they can summarize environmental feedback into natural language-based reflections and use these reflections to improve performance in subsequent trials. Self-Refine~\citep{madaan2024selfrefine} processes results from environmental interactions, using these outcomes to improve performance. Reflexion~\citep{shinn2024reflexion} involves introspection about feedback from the environment, generating reflective experiences that enhance reasoning abilities. Retroformer~\citep{yao2023retroformer} introduces a framework by learning a retrospective model to enhance LLM-based agents, which automatically adjusts language agent prompts based on environmental feedback through policy gradient. Nonetheless, these approaches typically rely on large-scale LLMs, resulting in significant costs and delays. Moreover, most smaller LLMs typically exhibit insufficient performance and robustness when deployed as agents.

\textbf{Language Agent Fine-tuning}~
To address the reliance of agent tasks on large-scale LLMs, the agent-tuning method is a standard solution. Researchers leverage powerful closed-source LLMs or human experts to generate expert trajectories, acting as a dataset for fine-tuning smaller open-source LLMs through imitation learning~\citep{ho2022large, chen2023fireact, zeng2023agenttuning, gou2023tora, yang2024react}. 
%Moreover, AutoAct~\citep{qiao2024autoact} enhances multi-agent systems by self-synthesizing trajectories but cannot reflect and learn from failures.
Moreover, AutoAct~\citep{qiao2024autoact} enhances the planning, reflection and action capabilities of multi-agent systems through self-synthetic trajectories. However, its reflection is only a summary of the current situation and does not have the ability to continue learning and evolution.
Compared to imitation learning, reinforcement learning aims to learn through self-exploration and trial-and-error without relying on the guidance of external experts. Archer~\citep{xi2024training} employs a hierarchical RL approach with two parallel RL algorithms to enhance the planning ability of LLM-based agents.
Several other works \citep{lightman2023let,chen2024RLMEC, havrilla2024teaching} propose intricate reward models with more fine-grained reward signals to improve performance.
In summary, these methods focus on training planner models through fine-tuning, but the fine-tuned agents usually struggle to retain their capability for learning and adaptation to the environment.

In summary, while these methods enhance planning models through fine-tuning, this process often diminishes the agent's ability for self-reflection.
%In Table~\ref{sample-table}, we summarize the recent language agent research related to our work from two dimensions and three features, distinguishing our method from others. The \textit{planner} refers to models that directly interact with the environment and make specific plans. In contrast, a \textit{reflector} refers to models that assist the planner in continuously reflecting on and evolving from mistakes. It can be seen that most methods currently focus on fine-tuning the planner component alone~\citep{chen2023fireact, zhou2024archer}, whereas Retroformer~\citep{yao2023retroformer} fine-tunes the reflector. However, no agent framework jointly optimizes the planner and reflector through a combination of supervised learning and reinforcement learning.

\section{More Detailed Experimental Settings}
\label{app: exp details}
\subsection{Evaluation Detials}
For HotpotQA and InterCode, we select 100 tasks as the test environments. For ALFWorld, we use 134 ``out-of-distribution'' tasks as the test environments. In all environments, we conduct ten rounds of trials and reflections. In terms of evaluation metrics, we report the \textit{F1 Score} for HotpotQA, the \textit{Success Rate} for ALFWorld, and the \textit{Reward Score (IoU)} for InterCode in the main text. Additionally, we provide the \textit{Exact Match Score} for HotpotQA and the \textit{Success Rate} data for InterCode as supplementary experiments in the Appendix \ref{app: more exp}. In tables, we report the initial reward (\textbf{IR}) to measure the planner's performance on the first trial, as well as the final reward (\textbf{FR}) and average reward (\textbf{AR}) to evaluate the performance of both the planner and reflector comprehensively.

\subsection{Baselines}

\quad\textbf{ReAct}~\citep{yao2023react}~is a method that combines reasoning and action in language models, enhancing performance in understanding and decision-making tasks by alternately generating reasoning trajectories and task-specific actions.

\textbf{Reflexion}~\citep{shinn2024reflexion}~builds on the ReAct framework, allowing language agents to learn from past errors by converting feedback into textual summaries, providing context for future tasks to improve performance through self-reflection.

\textbf{SFT}~\citep{chen2023fireact} is a fine-tuning method for language agents that enhances performance by utilizing expert trajectories, significantly improving efficiency and accuracy compared to prompt-based models. \textbf{We use the FireAct method to fine-tune the language agent separately on each dataset}.

\textbf{EI}~\citep{havrilla2024teachinglargelanguagemodels}(Expert Iteration) is a strong baseline that involves using a model initialized with SFT to generate data, filtering successful samples, and then using them to further fine-tune the model
%\textbf{Reflexion-IL}is a baseline algorithm designed to optimize both the planner and the reflector using imitation learning.

\textbf{RL}~We design an \textit{off-policy} policy gradient algorithm to fine-tune the planner model as a baseline. This algorithm follows the same reinforcement learning (RL) framework as used in \ourmodel, incorporating \textit{off-policy} RL with imitation learning regularization. A detailed comparison between this algorithm and the standard PPO algorithm is provided in Appendix \ref{app:ppo}, further justifying our choice of this baseline.

\subsection{Expert Dataset}
\begin{table*}[htbp]
\centering
\renewcommand{\arraystretch}{1.2} % Increase row spacing
\begin{tabular}{ccc}
  \toprule
  \textbf{Environment and Dataset} & \textbf{Positive Examples (Planner)} & \textbf{Positive Examples (Reflector)} \\
  \midrule
  HotpotQA & 6956 & 1304 \\
  ALFWorld & 693 & 221 \\
  InterCode & 866 & 97 \\
  \bottomrule
\end{tabular}
\caption{Imitation Learning Dataset}
\label{tab:il data}
\end{table*}
In our research, we collect an imitation learning dataset using expert models. The dataset encompasses three different environments. We employ our custom evaluator to retain the positive examples from this dataset. Detailed information is provided in Table \ref{tab:il data}.
%and the dataset is publicly available at: \url{https://github.com/XueyangFeng/RetroAct}.

\subsection{Prompt Details}
Across three datasets, Our prompt design on \textbf{HotpotQA} and \textbf{ALFWorld} follow \citep{yao2023react,shinn2024reflexion}, given that the aforementioned approach has not been previously tested on \textbf{InterCode}, we proceed by designing our prompts in accordance with the underlying design philosophy of the method described above.
\label{app: prompt}
\begin{mybox}[prompt:planner_prompt]{Intercode Few-shot Planner Generation}

\textcolor{blue}{\#Problem description}\vspace{0.3em}

Interact with a MySQL Database system using SQL queries to answer a question.\vspace{0.3em}

Use interleaving Thought, Action, Observation steps.\vspace{0.3em}

Thought can reason about the current situation, and Action can be two types:\vspace{0.3em}

(1) execute, which executes SQL code on the MySQL Database system\vspace{0.3em}

(2) submit, which indicates that the previous observation is the answer\\ \hspace*{\fill} \\

\textcolor{blue}{\#Few-shot description}
\vspace{0.3em}

Here are some examples.\vspace{0.3em}

\textbf{Question:}\textbf{ What are the names and grades for each high schooler?}\vspace{1em}

\textbf{Thought 1:} I should write a SQL command that selects the name and grade fields from a table about high schoolers.\vspace{0.3em}

\textbf{Action 1:}  execute [ SELECT\_ name, grade FROM high\_schoolers ]
\vspace{0.3em}

\textbf{Observation 1:}  Error executing query: Table ``network\_1.high\_schoolers'' doesn't exist\vspace{0.3em}

\begin{center}
......
\end{center}\vspace{1em}

\textbf{Thought x:} ...This should be the answer.\vspace{0.3em}

\textbf{Action x}: submit\vspace{0.5em}\\ \hspace*{\fill} \\

\textcolor{blue}{\#Target task information}\vspace{0.3em}

Here is the task:\vspace{0.3em}

\textbf{Question:} \textbf{What are the names of poker players, ordered ascending by the number of final tables they have made?}\vspace{0.3em}

\textbf{Thought 1:}

\end{mybox}

\begin{mybox}[prompt:reflector_prompt]{Intercode Few-shot Reflector Generation}

\textcolor{blue}{\#Problem description}\vspace{0.3em}

You will be given the history of a past experience in which you were placed in an environment and given a programming task to complete. You were unsuccessful in completing the task. Do not summarize your environment, but rather think about the strategy and path you took to attempt to complete the task. Devise a concise, new plan of action that accounts for your mistake with reference to specific actions that you should have taken. For example, if you tried A and B but forgot C, then devise a plan to achieve C with environment-specific actions. You will need this later when you are solving the same task. Give your plan after ``Plan''. \vspace{0.3em}\\ \hspace*{\fill} \\

\textcolor{blue}{\#Few-shot description}
\vspace{0.3em}

\textbf{Here are some examples:}\vspace{0.3em}

\textcolor{gray}{ [ \# the history of the failure trajectory  }\vspace{0.5em}

\textcolor{gray}{Question:} \hspace{2em}  \vspace{0.5em}

\textcolor{gray}{Thought x:} \hspace{1.5em}   \vspace{0.5em}

\textcolor{gray}{Action x:} \hspace{2em} 
\vspace{0.5em}

\textcolor{gray}{Observation x: }  \vspace{0.5em}

\begin{center}
\textcolor{gray}{......}
\end{center}\vspace{1em}

\textcolor{gray}{ ]}\vspace{0.5em}

\textcolor{red}{STATUS: FAIL} \vspace{0.5em}

\textcolor{orange}{New plan:}  \textcolor{black}{I was unsuccessful in finding the most common bond type because I did not consider the possibility of null or empty values in the ``bond\_type'' column. In the future, I should modify my SQL query to exclude these values. The revised plan would be to first check the tables in the database, then check the columns in the ``bond'' table, and finally execute a SQL query to find the most common bond type, excluding null or empty values. The new SQL query would be: ...} \\ \hspace*{\fill} \\

\textcolor{blue}{\#Target task information}\vspace{0.3em}

\textbf{Here is the task:}\vspace{0.3em}

\textcolor{gray}{ [ \# A trajectory that requires reflection  }\vspace{0.5em}

\textcolor{gray}{Question:} \hspace{2em}  \vspace{0.5em}

\textcolor{gray}{Thought x:} \hspace{1.5em}   \vspace{0.5em}

\textcolor{gray}{Action x:} \hspace{2em} 
\vspace{0.5em}

\textcolor{gray}{Observation x: }  \vspace{0.5em}

\begin{center}
\textcolor{gray}{......}
\end{center}\vspace{1em}

\textcolor{gray}{STATUS: FAIL }  \vspace{0.5em}

\textcolor{gray}{ ]}\vspace{0.5em}

\textcolor{purple}{Plans from past attempts:}\vspace{0.5em} \\ \hspace{8em} \#\#\#\hspace{1em}reflection of the past failure trajectory in this task\hspace{1em}\#\#\# \\

\textcolor{orange}{New plan:}

\end{mybox}

\subsection{Evaluation Metrics and Reward Function}
\label{sec:exp_details}
\label{app:eva}

We summarize the reward functions and evaluation metrics for the three datasets in a table \ref{tab:eva}. Additionally, we provide the final evaluation results for all these metrics.
\begin{table*}[htbp]
\centering
\renewcommand{\arraystretch}{1.5} % Increase row spacing
\begin{tabular}{ccc}
  \toprule
  \textbf{Environment and Dataset} & \textbf{Evaluator} & \textbf{Planner Reward Function} \\
  \midrule
  HotpotQA & EM Score (Eq.~\ref{eq: hotpotqa_em}) & F1 Score (Eq.~\ref{eq: hotpotqa_f1}) \\
  ALFWorld & \multicolumn{2}{c}{Success Rate (Eq.~\ref{eq: alfworld_sr})} \\
  InterCode & Success Rate (Eq.~\ref{eq: intercode_sr}) & Reward Score (Eq.~\ref{eq: intercode_reward}) \\
  \bottomrule
\end{tabular}
\caption{Evalator and Reward Function}
\label{tab:eva}
\end{table*}

\textbf{HotpotQA}~We employ two primary metrics to assess the performance of models: \textbf{Exact Match (EM)} and the \textbf{F1 Score}. The EM score is a strict metric that measures whether the normalized predicted answer exactly matches the normalized gold answer. It is defined as:
\begin{equation}
\label{eq: hotpotqa_em}
EM = 
\begin{cases} 
1 & \text{if prediction = gold} \\
0 & \text{otherwise}
\end{cases}
\end{equation}
The F1 Score is the harmonic mean of precision and recall. Precision (\(\text{precision}\)), recall (\(\text{recall}\)), and the F1 score (\(F1\)) are calculated as follows:
\begin{align}
\text{precision} &= \frac{\text{num\_same}}{\text{prediction}}, \quad \text{recall} = \frac{\text{num\_same}}{\text{gold}} \notag \\
F1 &= 2 \times \frac{\text{precision} \times \text{recall}}{\text{precision} + \text{recall}}
\end{align}

These metrics effectively evaluate the accuracy and reliability of answers generated by models in the HotpotQA dataset. When building the expert data set for imitation learning, we use the EM metric to filter the expert data. When performing reinforcement learning, we use the F1 score as the planner’s reward.

\textbf{ALFWorld}~Since ALFWorld only provides information on failure and success, we use whether the agent completes the task in the environment as the evaluation criterion and reward function.
\begin{equation}
\label{eq: alfworld_sr}
SR = 
\begin{cases} 
1 & \text{if agent completes the task} \\
0 & \text{otherwise}
\end{cases}
\end{equation}

\textbf{InterCode}~We employ two primary metrics to assess the performance of models: \textbf{Success Rate (SR)} and the \textbf{Reward Score}. The SR score is a strict metric that measures whether the SQL operations completed according to the task requirements. It is defined as:
\begin{equation}
\label{eq: intercode_sr}
SR = 
\begin{cases} 
1 & \text{if the result matches the gold answers} \\
0 & \text{otherwise}
\end{cases}
\end{equation}

The execution outcome of all SQL queries is a list of records. In order to more accurately and meticulously evaluate the results of the agent's command execution, we employ the same method as~\citep{yang2023intercode}, utilizing Intersection over Union (IoU), or more formally the Jaccard Index, to quantify the accuracy of the latest output generated by the agent in comparison to the gold standard output. Given the agent's latest execution output \(A\) and the gold answer's execution output \(G\), the reward function is as follows:
\begin{equation}
\label{eq: intercode_reward}
    \mathcal{R} = \frac{A \cap G}{A \cup G} \times \left(\frac{\text{kendalltau}(A \cap G, G \cap A) + 1}{2}\right)
\end{equation}

\subsection{Training Details}
\label{app: training_details}

~~~\textbf{Model Details}~We use Llama-chat-7b and Llama-chat-13b models in our experiments respectively. During training, we do not introduce additional prompt templates. Moreover, although calculating loss only in the model-generated parts of the trajectory is the optimal choice, splitting the trajectory significantly increases training costs. To balance training costs, we directly use the entire trajectory for auto-regressive training. We find that this approach does not result in significant performance loss while reducing training costs. 

\textbf{Inference Details}~During testing, we fix the \textit{temperature} parameter of all models to 0.0. This eliminates any randomness in the local models, ensuring the reproducibility of experiments and confirming that improvements in reflection are not due to randomness. In reinforcement learning training, we set the \textit{temperature} to 1.0 to allow for exploration. 
We use the \textit{vllm}~\citep{kwon2023efficient} framework to accelerate all inference processes. For HotpotQA, we limit the number of steps for a single trial to 5; for ALFWorld, we set the limit to 50; and for InterCode, we set the limit to 10.

\textbf{Training Details}~We implement LoRA based on PEFT~\cite{hu2021lora} and set $r_{LoRA} = 8$ and $\alpha_{LoRA} = 16$ for training in all experiments. We implement our own off-policy reinforcement learning algorithm based on the transformers~\citep{wolf-etal-2020-transformers}.

\textbf{System Specifications}~The system specifications for our experiments is shown in Table \ref{tab:system_specs}.
\label{app: system_specs}
\begin{table}[t]
    \centering
    \begin{tabular}{ll}
        \toprule
        \textbf{Name} & \textbf{Details} \\
        \midrule
        CPU     & Intel(R) Xeon(R) 8375C CPU \\
        GPU     & 4 * Nvidia A800 80GB PCIE \\
        Memory  & 1TB RAM \\
        Python  & Version 3.9 \\
        Transformers    & Version 4.38.2  \\
        \bottomrule
    \end{tabular}
    \caption{System Specifications}
    \label{tab:system_specs}
\end{table}

\textbf{Hyperparameters}~We set the regularization coefficient in RL, $\lambda_{planner}=1.0$,  $\lambda_{reflector}=1.0$ and the reward coefficient for the reflector, $\alpha = 1.0 $. Additionally, we conduct hyperparameter searches for the baseline IL and RL methods, as well as our \ourmodel method, within the range of learning rates $\{5e-05, 1e-04, 3e-04\}$ and epochs $\{3, 5\}$. In our experiments, due to the small data volume in the ALFWorld and InterCode datasets, we find that using a large batch size leads to insufficient update steps, resulting in severe underfitting. To maintain uniform settings, we use a batch size of 1 across all datasets. However, we strongly recommend increasing the batch size in scenarios with sufficient data volume to improve training stability.

\clearpage

\section{More Experiments}
\label{app: more exp}
\subsection{Comparison of Standard PPO and our proposed Off-policy policy gradient optimization}
\label{app:ppo}
The standard PPO algorithm for LLM was originally designed for standard Reinforcement Learning with Human Feedback (RLHF) tasks, particularly suitable for preference optimization scenarios. In such cases, since it's not feasible to obtain real-time human feedback during training, it relies on pre-labeled preference datasets to train a reward model, which then guides the reinforcement learning process. However, in LLM-Agent tasks, the agent interacts with the environment multiple times and can directly obtain rewards from it. Additionally, training a reward model in multi-step tasks is particularly challenging. We conduct an initial evaluation of the standard PPO algorithm's effectiveness in agent tuning tasks using the HotpotQA dataset:
\begin{figure}[h]
    \centering
    \includegraphics[width=0.5\textwidth]{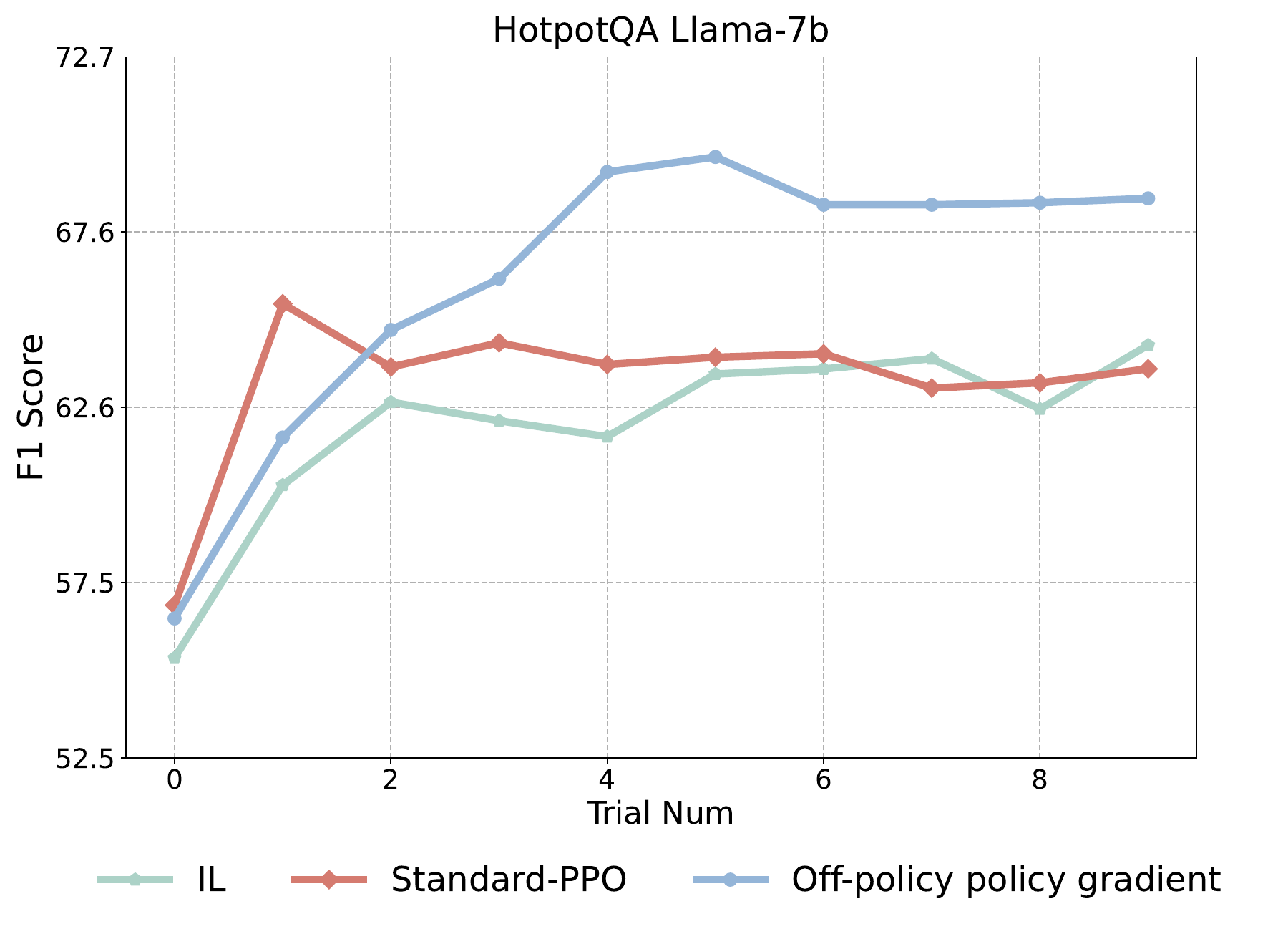}
    \caption{Experiments on Multi-Agent (Complete Results)}
     \label{fig:ppo}
\end{figure}

In the figure \ref{fig:ppo}, the performance of the standard PPO algorithm and imitation learning is similar, both lower than the Off-policy policy gradient algorithm we designed. In fact, we find that the standard PPO can easily cause the agent to lose the ability to follow instructions and fail to complete the task correctly. 
We analyze the reasons as follows: Standard PPO for RLHF~\citep{vonwerra2022trl} operates on a token-level reward mechanism, where the environment reward (provided by the reward model) only applies to the last token of a sequence, while the intermediate tokens rely on rewards provided by the continually learning critic model. Therefore, to achieve the desired results with PPO, typically, extensive training time and a large amount of data are required, along with thorough training of the reward model, critic model, and policy model to ensure proper convergence of the final objective. However, standard agent tasks often struggle to provide the high-quality data needed for this process. 

The standard token-level PPO algorithms, while theoretically capable of providing more fine-grained supervision signals, face significant challenges in achieving good convergence in agent tuning tasks. As demonstrated in~\citep{song2024trialerrorexplorationbasedtrajectory}, PPO, when used as a baseline, exhibit poor performance, making it a less meaningful and costly choice as a general RL baseline. Therefore, to make the experimental comparison more valuable, we use our designed off-policy policy gradient algorithm with imitation learning regularization as a more challenging RL baseline in our paper.

\subsection{Main Experiments on Multi-Agent (Complete Results and Additional Metrics)}
\label{app: multi exp}
\begin{figure*}[h]
    \centering
    \includegraphics[width=\textwidth]{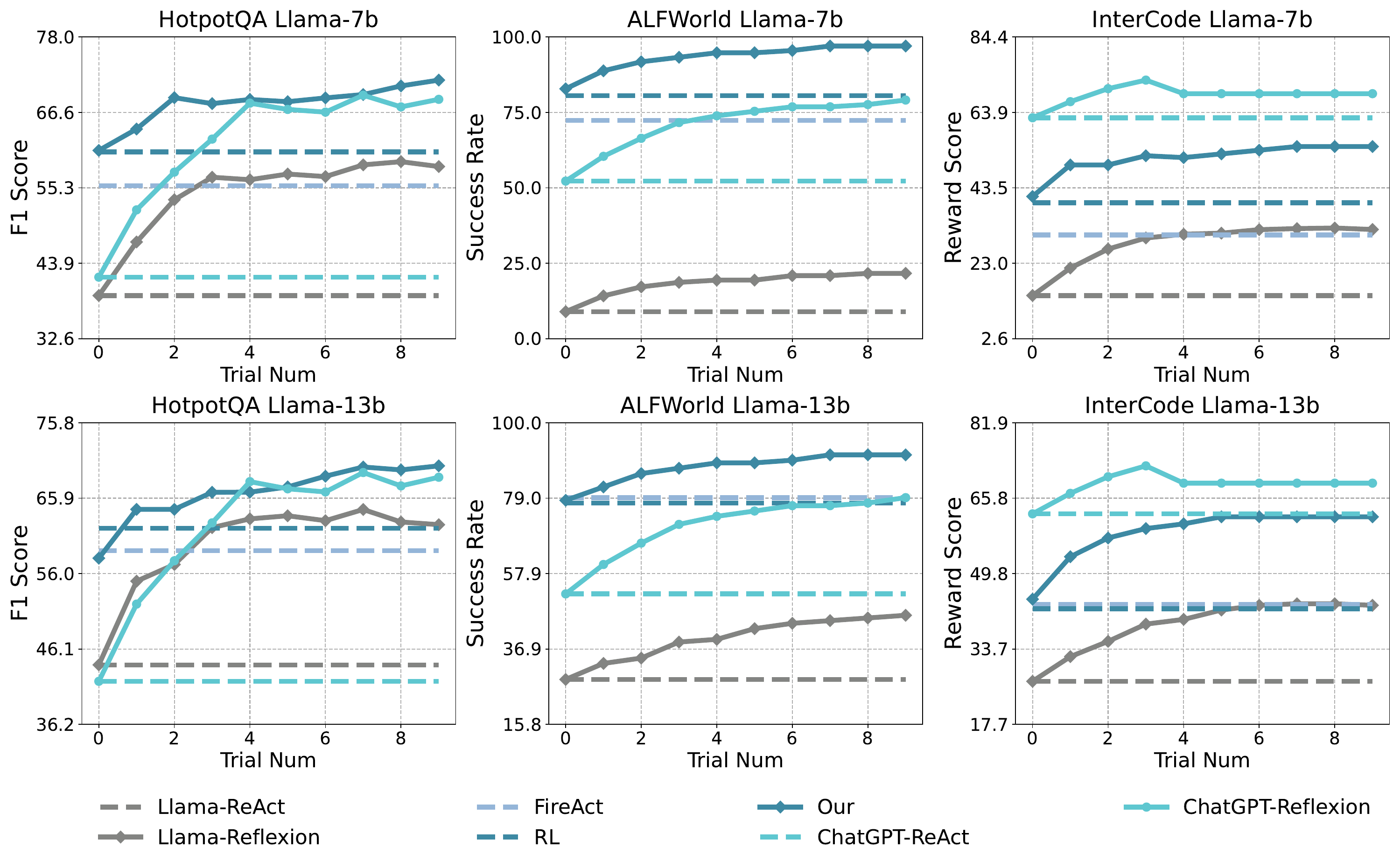}
    \caption{Experiments on Multi-Agent (Complete Results)}
     \label{fig:q1_app}
\end{figure*}

% \begin{figure}[h]
%     \centering
%     \includegraphics[width=\textwidth]{figures/q1_combined_2.pdf}
%     \caption{Experiments on Multi-Agent (Additional Metrics)}
%      \label{fig:q1_app2}
% \end{figure}

In this section, we present the complete data of our method and baseline methods in Figure \ref{fig:q1_app}. It can be observed that, whether in terms of rewards or additional metrics, the overall trends are consistent with the conclusions drawn in the main text. Our method outperforms the baseline methods significantly and is comparable to methods based on closed-source models.

\subsection{Error Analysis in ALFWorld}
\label{app:Classification of AlfWorld}
\begin{figure}[h]
    \centering
    \includegraphics[width=0.5\textwidth]{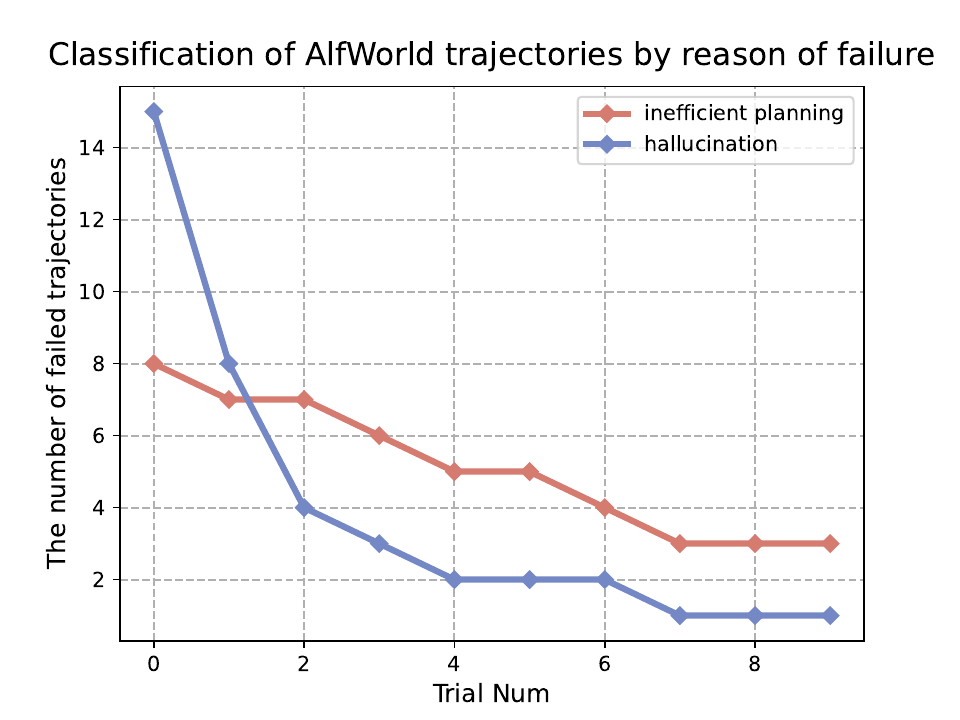}
    \caption{Error Analysis in ALFWorld}
     \label{fig:Classification of AlfWorld}
\end{figure}

\ourmodel significantly outperforms the baseline agent by completing 130 out of 134 tasks based on Llama-7b. In this section, we present a classification of ALFWorld trajectories by reason of failure in Figure \ref{fig:Classification of AlfWorld}, following \citep{shinn2024reflexion}. The reasons can be summarized into two points: \textbf{Hallucination}: The agent attempts to pick up non-existent items at a location and holds onto imaginary objects; \textbf{Inefficient planning}: The agent formulates plans that lack common sense and fails to make accurate judgments based on environmental feedback.

By analyzing the original trajectories, we find that fine-tuning and integrating prior knowledge effectively improves the rationality of the agent's actions, allowing it to track the placement of objects better. This underscores the crucial role of fine-tuning in enhancing the agent's capabilities. Moreover, by summarizing experiences, adjusting plans, and attempting multiple iterations, \ourmodel completes most of the previously failed tasks. This process highlights the essential role the self-reflection plays in improving \ourmodel's planning.

\subsection{Case Study}
\label{app: case}
In this section, we conduct a detailed and in-depth case study of our model. We deliberately select complex tasks from the original data that require multiple reflections to succeed. These cases comprehensively verify that our method can simultaneously enhance the model's planning and reflection capabilities.

\textbf{HotpotQA}~In HotpotQA, we deliberately select a challenging task requiring multi-step complex reasoning. The agent is tasked with answering the question: ``When Copsi was made Earl of Northumbria, he went back to reside in a town at the confluence of which two rivers?'' To answer this task, the agent needs to follow the correct reasoning process: \textbf{(1)} first, use a tool to search for Copsi and obtain information about his life; \textbf{(2)} then, correctly extract the town where Copsi resided as Earl of Northumbria from the returned information, avoiding other confusing details; \textbf{(3)} clearly understand that the question asks for the rivers at the town's location, not just the city itself; \textbf{(4)} correctly use a tool to search for the rivers at the town's location and answer the question accurately. We provide a specific case, comparing our agent (\ref{box:hotpotqa_pos}) with the baseline agent  (\ref{box:hotpotqa_nag}).

In \ourmodel agent's first attempt, it completes steps \textbf{1} and \textbf{2} but directly answers with the town. After two reflections, our agent correctly understands step \textbf{3} and clarifies the question details. Finally, in a post-reflection attempt, it completes the step \textbf{4}. In contrast, the baseline agent correctly completes step \textbf{1} but repeatedly fails to extract the correct information in step \textbf{2}. Even after nine subsequent reflections and attempts, it remains stuck at step \textbf{2}, falling into a loop.

This case demonstrates that our planner is superior to the baseline, as it can complete step \textbf{2} on the first attempt. Our reflector is also more effective, identifying the critical error during reflection and completing step \textbf{3}, and then helping the planner complete the \textbf{4}.

\textbf{AlfWorld}~In Alfworld, we select a challenging task that necessitates a clear and adaptive planning approach in response to the environment. The agent is in the middle of a room, where a multitude of items is arranged scattered across different spots, to look at the bowl under the desk lamp. To complete this task, the agent needs to follow the correct planning and reasoning process: \textbf{(1)} The agent needs to devise a strategy first to take the bowl, then use the desklamp to examine the bowl, as opposed to finding the desklamp and mistakenly taking it away then carrying the desklamp to find the bowl. \textbf{(2)} The agent is tasked with locating the bowl in accordance with a pre-established plan. \textbf{(3)} The agent, with the bowl in possession, proceeds to find the desklamp, intending to utilize it to look at the bowl. We provide a specific case, comparing our agent (\ref{box:alfworld_pos}) with the baseline agent  (\ref{box:alfworld_nag}).

In the beginning, the \ourmodel agent successfully formulates a plan that involves locating the bowl first and then proceeds to find the desklamp with the bowl in possession in step \textbf{1}. However, some hallucination issues still arise in steps \textbf{2} and \textbf{3}, such as attempting to take non-existent items or failing to act upon seeing the targeted objects. After five reflections, the agent clearly recognized the sequence: first, to locate the bowl, and then, to find the desklamp. It further leverages past trajectory experiences to discern the positions of various objects within the environment. Upon discovering the desklamp, it adopts a strategic approach by noting its location and only afterward proceeds to search for the bowl, intending to utilize the desklamp effectively once the bowl is in hand. In contrast, the baseline agent is struggling with devising an accurate path that initially finds the bowl and then takes it to find a desklamp, despite multiple rounds of reflection, fails to identify and rectify its mistakes in step \textbf{1}. It continues to experience hallucinations more severe than \ourmodel, leading to persistent inefficiencies in planning and execution even after repeated introspection cycles in step \textbf{2} and \textbf{3}. 

This case demonstrates that our planner is superior to the baseline because it can complete step \textbf{1} on the first attempt. It demonstrates an appropriate utilization of learned strategies and prior knowledge integrated to avoid mistakenly taking the desklamp away. Meanwhile, our planner is designed to minimize hallucinations in step \textbf{2} to the greatest extent possible, ensuring it executes actions in line with its predefined plan. Given the extensive space to explore within the room, mistakes are inevitable, even with a clear plan in place. Consequently, the agent must also possess the ability to accurately reflect upon and correct errors arising from the current environmental conditions. Our reflector is also more effective, identifying the critical error during reflection and helping the planner complete the target in step \textbf{3}. This highlights a need for substantial improvement in its reflector and the robustness of its planner to overcome these challenges.

\onecolumn
\begin{bluebox}[label=box:hotpotqa_pos]{Better Planning and Reflection in HotpotQA}
\label{box:hotpotqa_pos}
\textbf{Trial 1:} \\

\textbf{Question:} \\
When Copsi was made earl of Northumbria, he went back to reside in a town at the confluence of which two rivers? \\
\\
Thought 1: I need to search Copsi and Northumbria, find the town where Copsi resided as earl of Northumbria.\\

Action 1: \textcolor{darkgreen}{Search[Copsi]} \\

Observation 1: Copsi (or Copsig; Old English: \"C\u014dpsige\" ) [died 1067] was a Northumbrian magnate in late Anglo-Saxon England.  He was a supporter of Tostig and was exiled along with him in 1065.  Copsi soon fled to Orkney (then a part of Norway).  The next year (1066), he joined Tostig at Sandwich in Kent with 17 ships.  Copsi survived Tostig's defeat at Stamford Bridge. When William the Conqueror prevailed at Hastings, he traveled, in March 1067, to pay William homage at Barking (where William was staying while his tower was being constructed in London).  In return, William made Copsi earl of Northumbria and sent him back to \textcolor{darkgreen}{York}.  Copsi's rule lasted a mere five weeks. At that time, he was murdered by Osulf, son of Eadulf III of the ancient Bernician family, which had historically governed the area from Bamburghproves at Newburn-upon-Tyne.  Osulf, however, only ruled as earl until that autumn, when an outlaw killed him he was tracking. \\

\textcolor{blue}{\#Agent successfully extracts the correct town information ``York'' from a large amount of text.}\vspace{0.3em}

Thought 2: Copsi was made earl of Northumbria and resided in \textcolor{darkgreen}{York}. \\

\textcolor{blue}{\#Agent correctly extracts the town name on the first trial, but the question asks, ``Which two rivers converge in that town?''.}\vspace{0.3em}

Action 2: \textcolor{red}{Finish[York]} \\

Observation 2:\textcolor{red}{Answer is INCORRECT}. \\

...\\

\textbf{Trial 3:} \\

\textcolor{blue}{\#Agent successfully reflects on the previous error, identifying the cause as not correctly responding to the question. In its reflection, it explicitly states, ``Then find the rivers it is located on''.}\vspace{0.3em} \\
\textbf{Reflections:} \\
I searched Copsi and Northumbria, then found the town where Copsi was made earl of Northumbria. \textcolor{darkgreen}{I should have searched the town where Copsi was made earl of Northumbria and found the rivers where it was located}. \\

\textbf{Question:}  \\
When Copsi was made earl of Northumbria, he went back to reside in a town at the confluence of which two rivers? \\

Thought 1: I need to search Copsi and Northumbria, find the town where Copsi was made earl of Northumbria, and then \textcolor{darkgreen}{find the rivers where it was located}. \\

Action 1: \textcolor{darkgreen}{Search[Copsi and Northumbria]}\\

Observation 1: Copsi (or Copsig; Old English: \"C\u014dpsige\" ) [died 1067] was a Northumbrian magnate in late Anglo-Saxon England.  He was a supporter of Tostig, and was exiled along with him in 1065.  Copsi soon fled to Orkney (then a part of Norway).  The next year (1066), he joined Tostig at Sandwich, in Kent, with 17 ships.  Copsi survived Tostig's defeat at Stamford Bridge, and when William the Conqueror prevailed at Hastings he travelled, in March 1067, to pay William homage at Barking (where William was staying while his tower was being constructed in London).  In return, William made Copsi earl of Northumbria and sent him back to York.  Copsi's rule lasted a mere five weeks, at which time he was murdered by Osulf, son of Eadulf III of the ancient Bernician family which had historically governed the area from Bamburgh, at Newburn-upon-Tyne.  Osulf, however, only ruled as earl until that autumn, when he was killed by an outlaw he was tracking. \\

Thought 2: Copsi was made earl of Northumbria at York. I need to find the \textcolor{darkgreen}{rivers where York is located}.\\

\textcolor{blue}{\#Agent correctly responds to its own reflection by searching for the relevant rivers after identifying the town information, and ultimately answers the question correctly.} \\
Action 2: \textcolor{darkgreen}{Search[York rivers]} \\

Observation 2: York ( ) is a historic walled city at the confluence of the rivers \textcolor{darkgreen}{Ouse and Foss} in North Yorkshire, England.  The municipality is the traditional county town of the historic county of Yorkshire to which it gives its name.  The city has a rich heritage and has provided the backdrop to major political events in England throughout much of its two millennia of existence.  The city offers a wealth of historic attractions, of which York Minster is the most prominent, and a variety of cultural and sporting activities making it a popular tourist destination for millions. \\

Thought 3: York is located at the confluence of the rivers Ouse and Foss.\\

Action 3: \textcolor{darkgreen}{Finish[Ouse and Foss]}\\

Observation 3: \textcolor{darkgreen}{Answer is CORRECT}.
\end{bluebox}

\begin{redbox}[label=box:hotpotqa nag]{Poor Planning and Reflection in HotpotQA}
\label{box:hotpotqa_nag}
\textbf{Trial 1:} \\

\textbf{Question:} \\
When Copsi was made earl of Northumbria he went back to reside in a town at the confluence of which two rivers? \\
\\
Thought 1: I need to search Copsi and the two rivers, find the town.\\

Action 1: \textcolor{darkgreen}{Search[Copsi]}\\

Observation 1: Copsi (or Copsig; Old English: \"C\u014dpsige\" ) [died 1067] was a Northumbrian magnate in late Anglo-Saxon England.  He was a supporter of Tostig, and was exiled along with him in 1065.  Copsi soon fled to Orkney (then a part of Norway).  The next year (1066), he joined Tostig at Sandwich, in Kent, with 17 ships.  Copsi survived Tostig's defeat at Stamford Bridge, and when William the Conqueror prevailed at Hastings he travelled, in March 1067, to pay William homage at Barking (where William was staying while his tower was being constructed in London).  In return, William made Copsi earl of Northumbria and sent him back to York.  Copsi's rule lasted a mere five weeks, at which time he was murdered by Osulf, son of Eadulf III of the ancient Bernician family which had historically governed the area from Bamburgh, at \textcolor{red}{Newburn-upon-Tyne}.  Osulf, however, only ruled as earl until that autumn, when he was killed by an outlaw he was tracking.\\

\textcolor{blue}{\#Despite retrieving the correct information, the agent incorrectly extracts the town information from the search results.} \\
Thought 2: The town at the confluence of the two rivers is \textcolor{red}{Newburn-upon-Tyne}.\\

Action 2: \textcolor{red}{Search[Newburn-upon-Tyne]}\\

Observation 2: Two Rivers High School is a public high school for students in grades 7 through 12 located in Ola, Arkansas, United States.  Two Rivers High School is administered by the Two Rivers School District.  The Two Rivers High School and school district serves 650 sqmi of rural communities including Ola, Plainview, Rover, and Casa.  Two Rivers High School is one of four high schools within Yell County and the sole high school administered by the Two Rivers School District.\\

Thought 3: Two Rivers High School is not the town at the confluence of the two rivers.\\

Action 3: \textcolor{red}{Finish[Newburn-upon-Tyne]}\\

Observation 3: \textcolor{red}{Answer is INCORRECT} \\

\textbf{Trial 2:} \\

\textcolor{blue}{\#The agent never reflects on the useful information, leading it to remain stuck in a continuous loop.} \\
\textbf{Reflections:} \\
\textcolor{red}{I searched Copsi and the two rivers, but I did not find the town. I should have searched for the name of the town in the observation}. \\

Thought 1: I need to search for the town where Copsi resided.\\

Action 1: \textcolor{red}{Search[town at the confluence of two rivers]}\\

Observation 1: The Two Rivers is a 7.0 mi river in Kittson County, northwestern Minnesota, in the United States.  Formed by the North Branch of the Two Rivers and the South Branch of the Two Rivers, it is a tributary of the Red River of the North, with its outflow traveling north through Lake Winnipeg and the Nelson River to Hudson Bay.\\

Thought 2: The town where Copsi resided is not the Two Rivers. I need to search for the correct town. \\

Action 2: \textcolor{darkgreen}{Search[Copsi]} \\

Observation 2: Copsi (or Copsig; Old English: \"C\u014dpsige\" ) [died 1067] was a Northumbrian magnate in late Anglo-Saxon England.  He was a supporter of Tostig, and was exiled along with him in 1065.  Copsi soon fled to Orkney (then a part of Norway).  The next year (1066), he joined Tostig at Sandwich, in Kent, with 17 ships.  Copsi survived Tostig's defeat at Stamford Bridge, and when William the Conqueror prevailed at Hastings he travelled, in March 1067, to pay William homage at Barking (where William was staying while his tower was being constructed in London).  In return, William made Copsi earl of Northumbria and sent him back to York.  Copsi's rule lasted a mere five weeks. At that time, he was murdered by Osulf, son of Eadulf III of the ancient Bernician family which had historically governed the area from Bamburgh, at Newburn-upon-Tyne.  Osulf, however, only ruled as earl until that autumn, when an outlaw killed him he was tracking.\\

Thought 3:  \textcolor{red}{The town where Copsi resided is Newburn-upon-Tyne}. I need to search for the correct answer.\\

Action 3: \textcolor{red}{Search[Newburn-upon-Tyne]}\\

Observation 3: Two Rivers High School is a public high school for students in grades 7 through 12 located in Ola, Arkansas, United States.  The Two Rivers School District administers two Rivers High School.  The Two Rivers High School and school district serves 650 sqmi of rural communities, including Ola, Plainview, Rover, and Casa.  Two Rivers High School is one of four high schools within Yell County and the sole high school administered by the Two Rivers School District.\\

Thought 4: Two Rivers High School is not the town where Copsi resided. I need to search for the correct answer.\\

Action 4: \textcolor{red}{Search[Copsi]}\\

Observation 4: Copsi (or Copsig; Old English: \"C\u014dpsige\" ) [died 1067] was a Northumbrian magnate in late Anglo-Saxon England.  He was a supporter of Tostig, and was exiled along with him in 1065.  Copsi soon fled to Orkney (then a part of Norway).  The next year (1066), he joined Tostig at Sandwich, in Kent, with 17 ships.  Copsi survived Tostig's defeat at Stamford Bridge, and when William the Conqueror prevailed at Hastings he travelled, in March 1067, to pay William homage at Barking (where William was staying while his tower was being constructed in London).  In return, William made Copsi earl of Northumbria and sent him back to York.  Copsi's rule lasted a mere five weeks, at which time he was murdered by Osulf, son of Eadulf III of the ancient Bernician family which had historically governed the area from Bamburgh, at Newburn-upon-Tyne.  Osulf, however, only ruled as earl until that autumn, when he was killed by an outlaw he was tracking.\\

Thought 5: The town where Copsi resided is Newburn-upon-Tyne. I need to search for the correct answer.\\

Action 5: \textcolor{red}{Search[Newburn-upon-Tyne]}\\

Observation 5: Two Rivers High School is a public high school for students in grades 7 through 12 located in Ola, Arkansas, United States.  Two Rivers High School is administered by the Two Rivers School District.  The Two Rivers High School and school district serves 650 sqmi of rural communities including Ola, Plainview, Rover, and Casa.  Two Rivers High School is one of four high schools within Yell County and the sole high school administered by the Two Rivers School District. \\

\textbf{Trial 3-10:} \\

\textcolor{red}{Agent falls into a Trial2-like dead loop until the maximum number of attempts was reached.}

\end{redbox}

\begin{bluebox}[label=box:AlfWorld)_pos]{Better Planning and Reflection in AlfWorld}
\label{box:alfworld_pos}

\textbf{Here is the task:}
You are in the middle of a room. Looking quickly around you, you see a bed 1, a desk 2, a desk 1, a drawer 6, a drawer 5, a drawer 4, a drawer 3, a drawer 2, a drawer 1, a garbagecan 1, a laundryhamper 1, a safe 1, a shelf 6, a shelf 5, a shelf 4, a shelf 3, a shelf 2, and a shelf 1.

\textbf{Your task is to:} look at bowl under the desklamp. \\
\\

\textbf{Trial 1:} \\
> think: To solve the task, I need to \textbf{find and take a bowl},\textbf{ then find and use a desklamp}. First I need to find a bowl. A bowl is more likely to appear in drawer (1-6), shelf (1-6), bed (1), garbagecan (1), laundryhamper (1), safe (1). I can check one by one, starting with drawer 1.\\> go to drawer 1\\The drawer 1 is closed.\\...\\> open drawer 6\\You open the drawer 6. The drawer 6 is open. In it, you see a book 2, and a pen 2.\\> go to shelf 1\\On the shelf 1, you see a pencil 2.\\...\\> go to shelf 6\\On the shelf 6, you see nothing.\\> think: Now I find a bowl (1). Next, I need to take it.\\> take bowl 1 from shelf 6\\Nothing happens.\\> think: I am stuck in a loop. I will try to execute a different action.\\\textcolor{blue}{\# The agent experiences a hallucination, prompting it to make the motion of picking up a bowl where none actually exists.}\vspace{0.3em}

> go to bed 1\\On the bed 1, you see a book 1, a laptop 1, a pillow 2, and a pillow 1.\\> go to desk 2\\On the desk 2, you see a bowl 2, a bowl 1, a cd 2, a cd 1, and a creditcard 3.\\> think: Now I find a bowl (2). Next, I need to take it.\\> take bowl 2 from desk 2\\\textcolor{darkgreen}{You pick up the bowl 2 from the desk 2.}\\> think: Now I take a bowl (2). Next, I need to \textbf{find a desklamp}. A desklamp is more likely to \textbf{appear in desk (1-2)}, bed (1), shelf (1-6), drawer (1-6), garbagecan (1), laundryhamper (1), safe (1). I can check one by one, starting with desk 1.\\> go to desk 1\\On the desk 1, you see a alarmclock 1, a \textcolor{darkgreen}{desklamp 1}, a mug 1, and a pencil 1.\\> go to desk 2\\On the desk 2, you see a bowl 1, a cd 2, a cd 1, and a creditcard 3.\\\textcolor{blue}{\# Despite the presence of a target object, a desklamp, the agent fails to devise a plan incorporating the use of the desklamp to accomplish its task.}\vspace{0.3em} 

> use desklamp 1\\Nothing happens.\\> think: I am stuck in a loop. I will try to execute a different action.\\> go to garbagecan 1\\On the garbagecan 1, you see nothing.\\...\\> go to shelf 3\\On the shelf 3, you see \textcolor{darkgreen}{a desklamp 1}, and a pencil 1.\\> go to shelf 4\\On the shelf 4, you see a alarmclock 2, and a pencil 3.\\...\\> go to shelf 6\\On the shelf 6, you see nothing.\\> think: Now I find a bowl (1). Next, I need to take it.\\> \textcolor{darkgreen}{take bowl 1 from shelf 6}\\Nothing happens.\\> think: I am stuck in a loop. I will try to execute a different action.\\> go to bed 1\\...\\\textcolor{blue}{\# The agent appears to have disregarded its prior planning, encountering a desklamp yet neglecting to utilize it, proceeding instead to attempt fetching a bowl again, ultimately leading to task failure.}\vspace{0.3em}

\textbf{STATUS: FAIL}\\

\textbf{Trial 6:} \\

\textbf{Reflections:} \\
\textcolor{red}{I was stuck in a loop in which I continually tried to take a bowl from shelf 6. I should have found a bowl first, then taken it. I will try to execute a different action if I am stuck in a loop again.}

\textcolor{red}{I was stuck in a loop in which I continually tried to use a desklamp. I should have found a bowl first, then taken it, then used a desklamp to light the bowl. I will try to execute a different action if I am stuck in a loop again.}

...\\

> think: To solve the task, I need to \textbf{find and take a bowl}, then \textbf{find and use a desklamp}. First I need to find a bowl. A bowl is more likely to appear in drawer (1-6), shelf (1-6), bed (1), garbagecan (1), laundryhamper (1), safe (1). I can check one by one, \textcolor{darkgreen}{starting with desk 1}.\\> go to desk 1\\On the desk 1, you see a alarmclock 1, a desklamp 1, a mug 1, and a pencil 1.\\> think: Now I \textcolor{darkgreen}{find a desklamp} (1).I will try to execute a different action.\\> go to desk 2\\On the desk 2, you see a bowl 2, a bowl 1, a cd 2, a cd 1, and a creditcard 3.\\> think: Now I find a bowl (2). Next, I need to take it.\\> \textcolor{darkgreen}{take bowl 2 from desk 2}\\You pick up the bowl 2 from the desk 2.\\> think: Now I take a bowl (2). Next, I need to find a desklamp. A desklamp is more likely to appear in desk (1), shelf (1-6), bed (1), garbagecan (1), laundryhamper (1), safe (1). I can check one by one, \textcolor{darkgreen}{starting with desk 1}.\\> \textcolor{darkgreen}{go to desk 1}\\On the desk 1, you see a alarmclock 1, \textcolor{darkgreen}{a desklamp 1}, a mug 1, and a pencil 1.\\> think: Now I find a desklamp (1). Next, I need to use it.\\> \textcolor{darkgreen}{use desklamp 1}\\You turn on the desklamp 1.\vspace{0.3em}

\textbf{STATUS: OK}

\end{bluebox}

\begin{redbox}[label=box:hotpotqa nag]{Poor Planning and Reflection in ALfowrld}
\label{box:alfworld_nag}

\textbf{Here is the task:}
You are in the middle of a room. Looking quickly around you, you see a bed 1, a desk 2, a desk 1, a drawer 6, a drawer 5, a drawer 4, a drawer 3, a drawer 2, a drawer 1, a garbagecan 1, a laundryhamper 1, a safe 1, a shelf 6, a shelf 5, a shelf 4, a shelf 3, a shelf 2, and a shelf 1.

\textbf{Your task is to:} look at bowl under the desklamp. \\
\\

\textbf{Trial 1:} 

> think: To solve the task, I need to \textcolor{darkgreen}{find and use a desklamp}. A desklamp is more likely to appear in desk (2), desk (1), drawer (6), drawer (5), drawer (4), drawer (3), drawer (2), drawer (1), garbagecan (1), laundryhamper (1), safe (1), shelf (6), shelf (5), shelf (4),\\> go to desk 1\\On the desk 1, you see \textcolor{darkgreen}{a desklamp 1}, a keychain 3, a mug 2, a pen 2, a pen 1, and a pencil 2.\\> go to desk 2\\On the desk 2, you see a alarmclock 1, a bowl 1, and a cd 3.\\> go to drawer 6\\The drawer 6 is closed.\\...\\> go to drawer 1\\The drawer 1 is closed.\\...\\> go to shelf 6\\On the shelf 6, you see nothing.\\...\\> go to shelf 3\\On the shelf 3, you see  \textcolor{darkgreen}{a desklamp 1}, a pen 1, and a pencil 2.\\...\\> go to bed 1\\On the bed 1, you see a book 1, a cellphone 1, a laptop 2, a laptop 1, a pillow 2, and a pillow 1.\\> go to desk 1\\On the desk 1, you see a desklamp 1, a keychain 3, a mug 2, a pen 2, a pen 1, and a pencil 2.\\> think: Now I find a desklamp (1). Next, I need to use it.\\> use desklamp 1\\You turn on the desklamp 1.\\> look at bowl under desklamp\\Nothing happens.\\> go back\\Nothing happens.\vspace{0.3em}

\textbf{STATUS: FAIL}\\

\textbf{Trial 3:} \\

\textcolor{blue}{\#The agent navigates the environment under flawed planning strategies.} \\
\textbf{Reflections:} \\
\textcolor{red}{In this environment, my past attempts were unsuccessful because I tried to execute the task in the wrong order. I will try to execute the task in the correct order this time. I will go to desk 1, find the desklamp, then look for the bowl under the desklamp.} 

\textcolor{red}{In this environment, my plan was to find a bowl then find and use a desklamp. However, the task says to look at bowl under the desklamp. I should have looked for the desklamp then looked for the bowl. I noticed that the desklamp was found on desk 1. In the next trial, I will go to desk 1, find the lamp, then look for the bowl under the desklamp.}

...\\

> think: To solve the task, I need to find and use a desklamp. A desklamp is more likely to appear in desk (2), desk (1), drawer (6), drawer (5), drawer (4), drawer (3), drawer (2), drawer (1), garbagecan (1), laundryhamper (1), safe (1), shelf (6), shelf (5), shelf (4),\\> go to desk 2\\On the desk 2, you see a alarmclock 1, \textcolor{darkgreen}{a bowl 1}, and a cd 3.\\> go to desk 1\\On the desk 1, you see \textcolor{darkgreen}{a desklamp 1}, a keychain 3, a mug 2, a pen 2, a pen 1, and a pencil 2.\\...\\> go to drawer 1\\The drawer 1 is closed.\\> open drawer 1\\You open the drawer 1. The drawer 1 is open. In it, you see a creditcard 1.\\> go to safe 1\\...\\The safe 1 is closed.\\> open safe 1\\You open the safe 1. The safe 1 is open. In it, you see a keychain 2, and a keychain 1.\\> go to shelf 3\\On the shelf 3, you see \textcolor{darkgreen}{a desklamp} 1, a pen 1, and a pencil 2.\\...\\
\textcolor{blue}{\#The agent never reflects on the useful information, leading it to remain stuck in a continuous loop.} \\> go to shelf 1\\On the shelf 1, you see a cellphone 3, and a creditcard 2.\\> go to desk 1\\On the desk 1, you see \textcolor{darkgreen}{a desklamp 1}, a keychain 3, a mug 2, a pen 2, a pen 1, and a pencil 2.\\> think: Now I find a desklamp (1). Next, I need to use it.\\> use desklamp 1\\You turn on the desklamp 1.\\> \textcolor{darkgreen}{look at bowl under desklamp}\vspace{0.3em}

\textbf{STATUS: FAIL}\\

%\textcolor{blue}{\#The agent never reflected on the useful information, leading it to %remain stuck in a continuous loop.} \\

\textbf{Trial 3-10:} 

\textcolor{red}{Agent falls into a Trial3-like dead loop until the maximum number of attempts was reached.}

\end{redbox}

\end{document}